\pdfoutput=1

\documentclass[11pt]{article}

\usepackage[final]{acl}

\usepackage{times}
\usepackage{latexsym}

\usepackage[T1]{fontenc}

\usepackage[utf8]{inputenc}

\usepackage{microtype}

\usepackage{inconsolata}

\usepackage{multirow}
\usepackage{booktabs}
\usepackage{graphicx}
\usepackage{amsmath}
\usepackage{amsfonts}
\usepackage{makecell}
\usepackage{hyperref}
\usepackage{stfloats}
\usepackage{tablefootnote}
\usepackage{algorithm}
\usepackage{algorithmic}

\usepackage{color}
\usepackage{soul}
\usepackage{makecell}
\usepackage{multicol}
\usepackage{supertabular}
\usepackage{listings}
\usepackage{arydshln}
\usepackage{tocloft}
\usepackage{adjustbox}

\definecolor{lightblue}{rgb}{.8,.8,1}

\title{CrossICL: Cross-Task In-Context Learning via Unsupervised Demonstration Transfer}

\author{Jinglong Gao\quad
Xiao Ding\footnotemark[1]\quad
Lingxiao Zou\quad
Bing Qin \quad Ting Liu \\
\normalsize{Research Center for Social Computing and Interactive Robotics}\\[-.05cm]
\normalsize{Harbin Institute of Technology, China}\\[-.05cm]
{\small\tt\{jlgao, xding, lxzou, qinb, tliu\}@ir.hit.edu.cn}\\[-.05cm]}

\begin{document}
\maketitle
\begin{abstract}

\renewcommand{\thefootnote}{\fnsymbol{footnote}}
\footnotetext[1]{Corresponding Author}

In-Context Learning (ICL) enhances the performance of large language models (LLMs) with demonstrations. However, obtaining these demonstrations primarily relies on manual effort. In most real-world scenarios, users are often unwilling or unable to provide such demonstrations.
Inspired by the human analogy, we explore a new ICL paradigm CrossICL to study how to utilize existing source task demonstrations in the ICL for target tasks, thereby obtaining reliable guidance without any additional manual effort.
To explore this, we first design a two-stage alignment strategy to mitigate the interference caused by gaps across tasks, as the foundation for our experimental exploration.
Based on it, we conduct comprehensive exploration of CrossICL, with 875 NLP tasks from the Super-NI benchmark and six types of LLMs, including GPT-4o. Experimental results demonstrate the effectiveness of CrossICL and provide valuable insights on questions like the criteria for selecting cross-task demonstrations, as well as the types of task-gap-induced interference in CrossICL.

\end{abstract}

\section{Introduction}

Recently, researchers improve the performance of large language models (LLMs) by providing demonstrations in the prompt \citep{brown2020language}, which is referred to as In-Context Learning (ICL).
However, ICL requires humans to provide demonstrations of the current task. In most real-world scenarios (such as accessing LLMs via apps or websites), users typically lack the ability or are unwilling to provide such demonstrations \citep{chen2023self,dong2024survey}, as it exceeds their capabilities and introduces additional manual effort. This hinders the practical application of ICL.

\begin{figure}[t]
  \centering
  \includegraphics[width=1\linewidth,trim={0cm 0cm 0.05cm 0cm},clip]{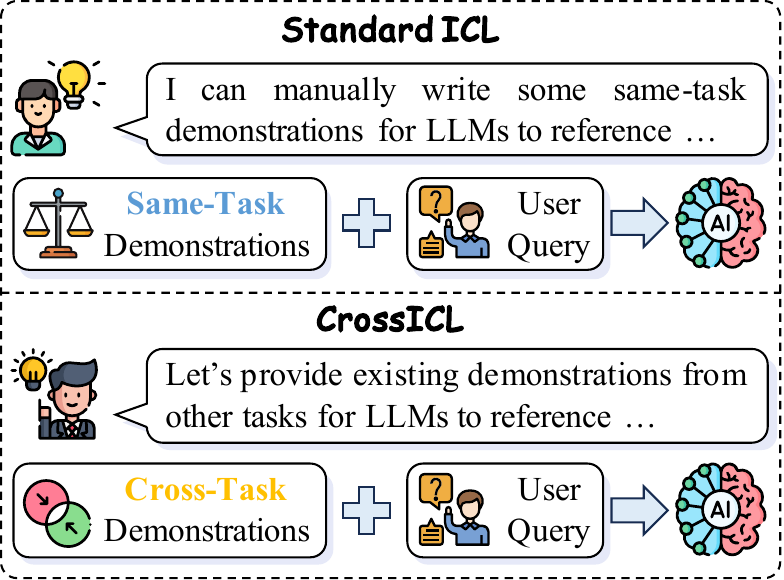}
  \caption{The standard ICL and our CrossICL, which performs ICL with existing cross-task demonstrations.}
  \label{fig:intro}
\end{figure}

To tackle this problem, previous studies explore zero-shot ICL, which allows ICL without human-provided demonstrations. SG-ICL \citep{kim2022self} guides LLMs to automatically complete task-specific demonstration templates, with a predefined set of question templates and output labels, introducing additional manual effort. Z-ICL \citep{lyu2023z} just retrieves sentences from external text corpora as new questions, and assigns random labels to them, resulting in noisy demonstrations. Both SG-ICL and Z-ICL are limited to several classification tasks. To better achieve zero-shot ICL, Self-ICL \citep{chen2023self} directly utilizes prompts to guide LLMs in generating diverse demonstrations based on the task description, making it applicable to various tasks. However, \textbf{due to the lack of guidance}, the quality of their generated demonstrations are not guaranteed, which can hurt the performance \citep{su-etal-2024-demonstration}.

The idea of ICL aligns with the human analogy \citep{dong2024survey}. However, the human analogy can benefit from different tasks \citep{GENTNER1983155}, while current ICL methods are limited to analogies within the same task. Inspired by this, we explore \textbf{a new ICL paradigm} \textbf{CrossICL} to study how to leverage cross-task demonstrations for ICL, applying existing source task demonstrations to new target tasks, thereby eliminating the need for manual efforts or unguided LLMs to synthesize demonstrations from scratch.
This new paradigm extends the in-task analogies of ICL to transfer learning with cross-task analogy, providing a new path for the generalization of LLMs and effectively broadening the application prospects of ICL.

However, ICL is highly sensitive to the differences between the queries and demonstrations \citep{min-etal-2022-rethinking}. Moreover, there are often gaps between the source and target tasks (e.g., task goals, input formats, output spaces, etc.), which can interfere with LLMs (\S\ref{sec:crosstaskerroranalyze}).

To mitigate this issue, we design a concise implementation of CrossICL as the foundation for our experimental exploration. Specifically, we narrow the source-target task gaps through a two-stage alignment strategy: \emph{1) Minimum Gap Selection}: For a given target task query, we first select the demonstrations from existing source task datasets that have the smallest gap to it.
As ICL typically assists LLMs in informing input-label mappings (as summarized by the task description) and input distributions \citep{min-etal-2022-rethinking,pan-etal-2023-context}, we primarily select based on these aspects.
\emph{2)~Progressive Task Adaptation}: For the selected demonstrations, we guide LLMs to autonomously transform them to further align with the target task. Specifically, LLMs first transform the source task demonstration query into a new target task query. Then, LLMs check and refine the new query. After that, LLMs generate the label of the new query with the guidance of the original source task demonstration.
As the final step, the aligned demonstrations are employed for ICL to answer the target task query.

Based on the above implementation, we conduct \textbf{comprehensive experimental exploration of CrossICL}, using 875 tasks from the Super-NI benchmark and six types of LLMs, including GPT-4o. Our main topics and findings are as follows:
1) We find that CrossICL effectively improves the performance of al six types of LLMs;
2) We conduct an in-depth analysis of how to select cross-task demonstrations from different source tasks, involving 7 key factors and 12 selection criteria;
3) We carefully study and summarize the interference caused by cross-task gaps in the reasoning processes of LLMs, ultimately identifying 7 major types of cross-task interference;
4) Experiments show that CrossICL still effectively improves model performance even with source tasks that are less similar to the target task;
5) Similar to standard ICL, both too many and too few demonstrations can decrease the performance of CrossICL;
6) Additionally, combining CrossICL with Query-Supervised ICL presents a potentially more powerful ICL method.

\section{Preliminary}
\label{sec:prel}
Following previous zero-shot ICL study \citep{chen2023self}, for the queries of a task, we represent them as a concatenation of two parts: 1)~\textbf{Task Description}: a relaxed statement of the task objective, e.g, ``Please choose the correct option'' or ``Give a title for the input''. It is mainly used for distinguishing tasks; 2) \textbf{Task Input}: The input to be processed, e.g., ``Question: ... Options: ...''.

For a user query $q_t$, it consists of the task description $d_t$ and the task input $x_t$ for the target task. The CrossICL selects $n$ existing source task demonstrations $\{s_1,\cdots,s_k,\cdots,s_n\}$, where the $k$-th demonstration $s_k$ consists of its task description $d_k$, task input $x_k$, and label $y_k$. Thus, CrossICL can be formalized as:
\begin{equation*}
 \hat{y}_t = \underset{y}{\mathrm{argmax}}\; p\left( y \mid s_1 \oplus \cdots \oplus s_n \oplus q_t, \theta \right) ,
\end{equation*}
where $\theta$ represents the parameters of the LLM. However, ICL is highly sensitive to the gap between the query and demonstrations \citep{min-etal-2022-rethinking,pan-etal-2023-context}, and therefore directly using source task demonstration can interfere with the inference of LLMs (\S\ref{sec:crosstaskerroranalyze}). The key to CrossICL lies in how to narrow such cross-task gap (e.g., task objectives, input formats, output spaces, etc.).

\section{Methodology}

\begin{figure*}[t]
  \centering
  \includegraphics[width=\linewidth,trim={0cm 0cm 0cm 0cm},clip]{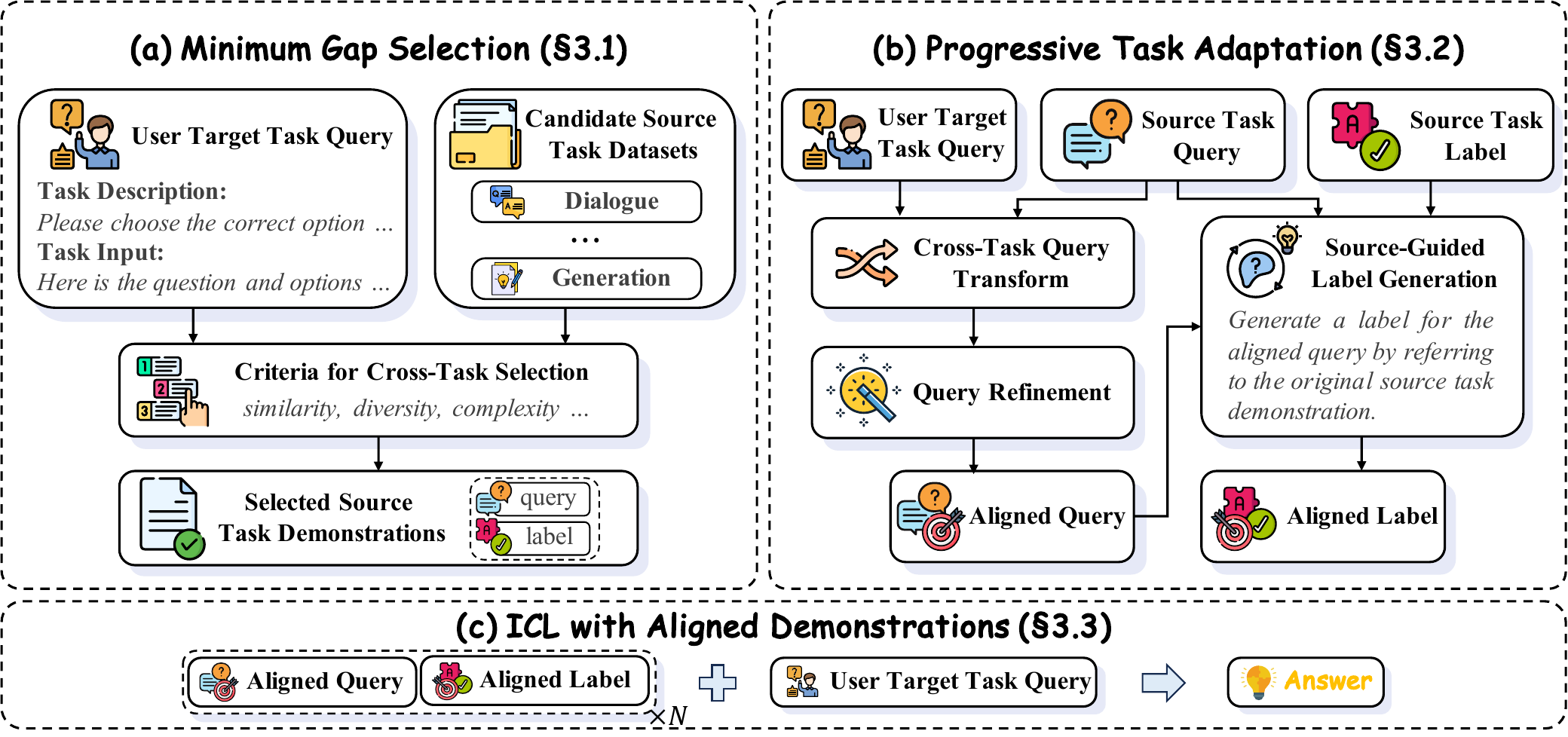}
  \caption{The framework of our proposed CrossICL.}
  \label{fig:framework}
\end{figure*}

As shown in Figure~\ref{fig:framework}, we propose a concise and universal CrossICL framework to mitigate the above issue, utilizing a two-stage alignment strategy to narrow the gap between source task demonstrations and the user target task query.
For each user query, we first perform Minimum Gap Selection, choosing the source task demonstrations with the smallest gap to the user query. Then, we conduct Progressive Task Adaptation, which involves multi-stage transformation to perform further alignment. Finally, we employ the aligned demonstrations for ICL to answer the initial user query.
Examples of our framework are shown in Appendix \ref{appendix:case_study_our}.

\subsection{Minimum Gap Selection}
\label{sec:memory}

To reduce the disturbance caused by the cross-task gap, we select source task demonstrations with the smallest gap to the target task query.

\subsubsection{How to Select ICL Demonstrations Across Tasks}
\label{sec:howtoselect}

For the minimum gap selection, the key issue is which criteria are employed to select the demonstrations from candidate source tasks.

According to previous studies \citep{min-etal-2022-rethinking,pan-etal-2023-context}, ICL primarily benefits LLMs by providing information across three aspects: 1)~input-label mapping, which is embedded in the task description, i.e., what kind of output is expected from the input; 2) input distributions, refers to the distribution of the queries in the task; 3) label space, the potential outputs of the task.
Therefore, large differences in these aspects will interfere with the ICL. The similarities in these aspects are suitable as criteria for minimum gap selection.
However, since the exact output for a user query is inaccessible, leading to unreliable selections, we only consider the first two aspects as selection criteria.

To better answer how to select cross-task demonstrations, we further provide an in-depth analysis in \S\ref{sec:diff_select}, including 7 factors and 12 selection criteria.

\subsubsection{Selection: Preparation and Inference}
Following the above analysis, we select the source task demonstrations based on the similarities of the task description and the task input .
The former gives a general description of the input-label mapping, while the latter clarifies the input distribution.

During the preparation, we encode all the source task datasets.
For the $k$-th source task demonstrations $s_k=\{q_k,y_k\}$, we employ an embedding model to encode its task description $d_k$ and its entire query $q_k=d_k \oplus x_k$ into vectors $\boldsymbol{v}_k^d$ and $\boldsymbol{v}_k^q$, respectively. Please note that demonstrations of the same task share the same task description, so $\boldsymbol{v}_k^d$ only needs to be computed once for each task.

During the inference phase, for a user query $q_t=d_t \oplus x_t$, we first encode its task description $d_t$ and the entire query $q_t$ into semantic vectors $\boldsymbol{v}_t^d$ and $\boldsymbol{v}_t^q$, respectively. Then, we identify the source task $\mathcal{S}$ whose description embedding $\boldsymbol{v}_\mathcal{S}^d$ has the highest cosine similarity to $\boldsymbol{v}_t^d$:
\begin{equation*}
    \hat{\mathcal{S}} = \underset{\mathcal{S}}{\mathrm{argmax}}\;  \cos(\boldsymbol{v}_\mathcal{S}^d, \boldsymbol{v}_t^d).
\end{equation*}
Then, we select top $n$ demonstrations of the identified source task $\hat{\mathcal{S}}$ that show the highest cosine similarity to $\boldsymbol{v}_t^q$:
\begin{equation*}
    \hat{s_k} = \underset{s_k\in D_\mathcal{S}}{\mathrm{argmax}}\;  \cos(\boldsymbol{v}_k^q, \boldsymbol{v}_t^q) .
\end{equation*}

\subsection{Progressive Task Adaptation}
\label{sec:adaptation}

To further narrow the cross-task gap, we guide LLMs to concurrently transform each selected source task demonstration into the target task.
Considering the complexity of one-step adaptation, we design a multi-step strategy to progressively achieve this alignment process. For each prompt, we use a concise design and avoid task-specific features, ensuring its generalizability and the representativeness of the evaluation results.
The prompts are shown in Appendix~\ref{appendix:prompts}.

\subsubsection{Cross-Task Query Transform}
For each source task demonstrations, we first transform its query to align with the target task query. This is necessary because LLMs are sensitive to the differences between demonstrations and queries.

Specifically, we use Prompt-\ref{pt:p1} to guide the LLMs in rewriting the source task query into a new target task query. The user target task query serves as an example for the LLMs to refer to.

Although LLMs can transform the query, due to the mixing of the source task and the target task in the prompt, the generated query often contains noise, such as format inconsistencies or the inclusion of source task keywords.

\subsubsection{Query Refinement}
To alleviate the noise, we further refine the synthesized query from the previous step.

Specifically, we use Prompt-\ref{pt:p2} to guide the LLMs in modifying the synthesized query based on the user target task query.

After this step, the query from the source task demonstration is transformed into a new query for the target task, but it still lacks the label needed to form a complete demonstration.

\subsubsection{Source-Guided Label Generation}

In this step, we generate the label for the refined query, forming a complete demonstration.

The original source task demonstration and its transformed query are highly similar.
Therefore, we use Prompt-\ref{pt:p3} to guide the LLMs in predicting the label for the synthesized query by referencing the original source task demonstration.
Since both usually involve similar topics, domains, and knowledge, this source-guided prompt improves the quality of label generation.

\subsection{ICL with Aligned Demonstrations}
Finally, we perform standard ICL with the source task demonstrations aligned through our selection-adaptation processes. Specifically, the aligned demonstrations are concatenated before the user query to guide the LLM in generating a reliable response (details in Prompt-\ref{pt:p4}).

\section{Experiments}
\subsection{Datasets and Evaluation Metrics}
\label{datasets}

We employ the Super-NI benchmark \citep{wang-etal-2022-super}, which is designed to evaluate the cross-task generalization ability of LLMs.
Following \citet{wang-etal-2024-inscl}, we utilizes 875 English tasks in Super-NI, including 756 training tasks and 119 test tasks, with the test tasks are divided into six categories. We only select source task from the training tasks.
For more details, please refer to Appendix~\ref{superni-more}.
Following the Super-NI setup, we employ the top 100 samples of each test task and utilize ROUGE-L \citep{lin2004rouge} as the evaluation metric.
The author of Super-NI carefully designed it to ensure that all tasks can be reliably evaluated using ROUGE-L. We also report accuracy metric in Appendix~\ref{app:em}. For GPT-4o, we randomly select 500 test samples from each category for evaluation due to its high cost.
We report the average score of three rounds of predictions to reduce randomness. Additionally, ``\textit{Avg.}'' refers to the average score across six categories.

\subsection{Parameters Setting}
\label{sec:paramsetting}
We evaluate CrossICL on six types of LLMs: Llama3.1-8B-Instruct \citep{dubey2024llama}, Gemma2-9B-It \citep{gemmateam2024gemma2improvingopen}, Qwen2-7B-Instruct \citep{yang2024qwen2technicalreport}, Qwen2.5-7B-Instruct \citep{qwen2025qwen25technicalreport}, Deepseek-7B-Chat \citep{deepseekai2024deepseekllmscalingopensource} and GPT-4o (\texttt{gpt-4o-2024-05-13}, \citet{openai2024gpt4technicalreport}).
The \texttt{temperature} of these LLMs are set to their default or commonly used values, which are 0.6, 0.6, 0.7, 0.7, 0.7, and 1, in order.
Among these LLMs, the first three are with similar performance. Compared to the first three, Qwen2.5-7B-Instruct offers stronger performance, while Deepseek-7B-Chat is relatively weaker. GPT-4o is one of the most powerful LLMs today. Given the high popularity of the Llama model, we mainly use it as the base LLM for analysis.
The number of source task demonstrations used for ICL in our framework is 5. The embedding model we employed is BGE-EN-ICL \citep{chen2024bgem3embeddingmultilingualmultifunctionality}.

\subsection{Baselines}
In our experiments, we employ the following baseline methods:
\underline{\textbf{Zero-Shot}}, let LLMs directly perform zero-shot inference for the user query;
\underline{\textbf{Zero-shot-CoT}} \citep{zscot}, add ``Let's think step by step'' after the query to prompt LLMs think with chain-of-thought (CoT);
\underline{\textbf{Self-ICL}}\footnote{\url{https://github.com/ntunlplab/Self-ICL}} \citep{chen2023self}, a zero-shot ICL method that guides LLMs to autonomously generate new demonstrations for ICL based on the given task description.
We also compare with the \textbf{demonstration-selection-based methods} in Appendix~\ref{app:diff_select_notransfer}.
Besides, we discuss the \textbf{query-supervised ICL} \citep{su-etal-2024-demonstration} in \S\ref{sec:qsicl}.

\begin{table*}[!t]
\small
\centering
\setlength{\tabcolsep}{3pt}
\begin{tabular}{llccccccc}
\toprule
\textbf{Model} & \textbf{Method} & \textbf{Classification} & \textbf{Comprehension} & \textbf{Dialogue} & \textbf{Extraction} &  \textbf{Generation} & \textbf{Rewriting} & \textit{\textbf{Avg.}} \\
\midrule
\multirow{4}{*}{\textbf{Llama3.1-8B}}
 & \textbf{Zero-Shot} & 0.602 & 0.512 & 0.655 & 0.507 & \textbf{0.398} & 0.520 & 0.532 \\
& \textbf{Zero-shot-CoT} & \textbf{0.640} & 0.524 & \underline{0.691} & 0.462 & 0.376 & 0.448 & 0.523 \\
& \textbf{Self-ICL} & 0.604 & \underline{0.532} & 0.654 & \underline{0.519} & \underline{0.388} & \underline{0.525} & \underline{0.537} \\
& \textbf{Ours} & \underline{0.635} & \textbf{0.566} & \textbf{0.693} & \textbf{0.539} & 0.383 & \textbf{0.593} & \textbf{0.568} \\
\midrule
\multirow{4}{*}{\textbf{Gemma2-9B}}
 & \textbf{Zero-Shot} & 0.608 & \underline{0.613} & 0.629 & \textbf{0.580} & \underline{0.404} & 0.507 & 0.557 \\
& \textbf{Zero-shot-CoT} & 0.610 & \textbf{0.646} & 0.608 & 0.531 & 0.380 & 0.411 & 0.531 \\
& \textbf{Self-ICL} & \underline{0.614} & 0.611 & \underline{0.644} & 0.574 & \textbf{0.404} & \underline{0.510} & \underline{0.560} \\
& \textbf{Ours} & \textbf{0.671} & 0.607 & \textbf{0.713} & \underline{0.578} & 0.400 & \textbf{0.593} & \textbf{0.594} \\
\midrule
\multirow{4}{*}{\textbf{Qwen2-7B}}
 & \textbf{Zero-Shot} & 0.692 & 0.579 & 0.681 & 0.477 & 0.387 & 0.386 & 0.534 \\
& \textbf{Zero-shot-CoT} & 0.603 & \underline{0.588} & 0.626 & 0.445 & 0.334 & 0.328 & 0.487 \\
& \textbf{Self-ICL} & \textbf{0.705} & 0.561 & \underline{0.686} & \underline{0.485} & \textbf{0.394} & \underline{0.400} & \underline{0.539} \\
& \textbf{Ours} & \underline{0.703} & \textbf{0.590} & \textbf{0.696} & \textbf{0.499} & \underline{0.393} & \textbf{0.519} & \textbf{0.567} \\
\midrule
\multirow{4}{*}{\textbf{Qwen2.5-7B}}
 & \textbf{Zero-Shot} & 0.546 & 0.547 & \underline{0.734} & \underline{0.597} & 0.397 & 0.513 & 0.555 \\
& \textbf{Zero-shot-CoT} & \underline{0.623} & \textbf{0.623} & 0.729 & 0.533 & 0.371 & 0.403 & 0.547 \\
& \textbf{Self-ICL} & 0.573 & 0.552 & 0.720 & 0.585 & \underline{0.405} & \underline{0.518} & \underline{0.559} \\
& \textbf{Ours} & \textbf{0.704} & \underline{0.577} & \textbf{0.741} & \textbf{0.610} & \textbf{0.416} & \textbf{0.596} & \textbf{0.607} \\
\midrule
\multirow{4}{*}{\textbf{Deepseek-7B}}
 & \textbf{Zero-Shot} & 0.512 & \underline{0.419} & \underline{0.522} & \underline{0.451} & 0.296 & 0.395 & 0.433 \\
& \textbf{Zero-shot-CoT} & 0.516 & \textbf{0.445} & 0.507 & 0.391 & 0.285 & 0.372 & 0.419 \\
& \textbf{Self-ICL} & \underline{0.523} & 0.415 & 0.501 & \textbf{0.461} & \underline{0.299} & \underline{0.416} & \underline{0.436} \\
& \textbf{Ours} & \textbf{0.534} & 0.418 & \textbf{0.535} & 0.428 & \textbf{0.311} & \textbf{0.469} & \textbf{0.449} \\
\midrule
\multirow{4}{*}{\textbf{GPT-4o}}
 & \textbf{Zero-Shot} & 0.745 & 0.744 & 0.782 & \underline{0.565} & \textbf{0.438} & \underline{0.593} & 0.644 \\
& \textbf{Zero-shot-CoT} & \underline{0.761} & \underline{0.779} & 0.780 & 0.552 & 0.429 & 0.507 & 0.634 \\
& \textbf{Self-ICL} & 0.756 & 0.757 & \underline{0.783} & \textbf{0.574} & 0.430 & 0.587 & \underline{0.648} \\
& \textbf{Ours} & \textbf{0.796} & \textbf{0.795} & \textbf{0.792} & 0.564 & \underline{0.434} & \textbf{0.672} & \textbf{0.676} \\
\bottomrule
\end{tabular}
\caption{\label{tab:main}Experimental results (\%) on the Super-NI benchmark. \textbf{Bold} and \underline{Underlined} represent the 1st and the 2nd best-performing methods for each LLM.
``\textit{Avg.}'' denotes the average score across six task categories.}
\end{table*}

\begin{figure*}[t]
  \centering
\includegraphics[width=1\linewidth,trim={0cm 0.2cm 0cm 0.1cm},clip]{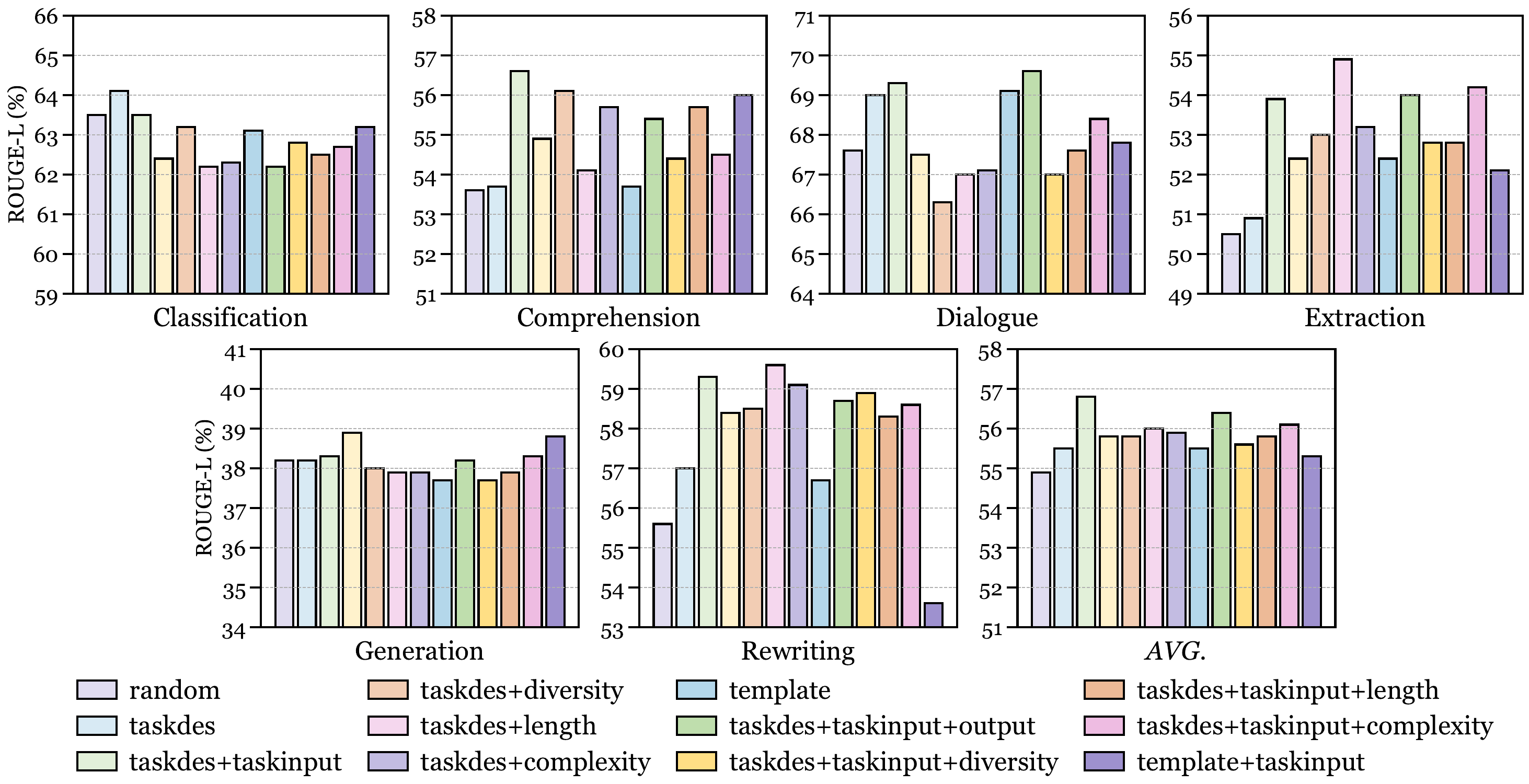}
  \caption{Performance (\%) of CrossICL based on Llama3.1-8B with different demonstration selection criteria.}
  \label{fig:diff_select_our}
\end{figure*}

\subsection{Main Results}

Table~\ref{tab:main} shows the experimental results on the Super-NI benchmark. Our findings are as follows:

Firstly, our CrossICL consistently outperforms the baseline methods.
This is mainly because CrossICL effectively leverages existing source task demonstrations in solving the target task, providing more reliable guidance for the LLMs.
Besides, we also discuss \textbf{the efficiency of ICL and our method} in Appendix~\ref{app:costcomp}.

Secondly, Self-ICL offers less benefits to LLMs, which is consistent with the conclusion of \citet{su-etal-2024-demonstration}. This is because Self-ICL completely relies on the LLMs own generation capability. Therefore, the generated poor demonstrations hurt the performance of ICL \citep{su-etal-2024-demonstration}. In contrast, our CrossICL provides more reliable guidance with existing source task demonstrations.

Thirdly, compared to the first three LLMs, CrossICL achieves a greater average improvement with Qwen2.5-7B. This is primarily because Qwen2.5-7B builds upon Qwen2 with more knowledge and task diversity, achieving better cross-task alignment of demonstrations.

Besides, although our CrossICL is still effective, the improvement on Deepseek-7B and GPT-4o is slightly smaller.
For Deepseek-7B, due to its relatively weaker capabilities, it struggles to align source task demonstrations with the target task, which hurts the performance of ICL. For GPT-4o, its high score in zero-shot inference makes the improvement from adding extra demonstrations less apparent compared to smaller LLMs.

\begin{table*}[!t]
\small
\centering
\setlength{\tabcolsep}{6pt}
\begin{tabular}{lccccccc}
\toprule
\textbf{Method} & \textbf{Classification} & \textbf{Comprehension} & \textbf{Dialogue} & \textbf{Extraction} &  \textbf{Generation} & \textbf{Rewriting} & \textit{\textbf{Avg.}} \\
\midrule
 \textbf{Ours} & \textbf{0.635} & \textbf{0.566} & \textbf{0.693} & \textbf{0.539} & \underline{0.383} & \textbf{0.593} & \textbf{0.568} \\
\textbf{- w/o Src in LG} & 0.629 & \underline{0.555} & \underline{0.671} & \underline{0.525} & 0.381 & 0.573 & \underline{0.556} \\
\textbf{- w/o Src in All} & 0.610 & 0.531 & 0.668 & 0.522 & 0.382 & 0.537 & 0.542 \\
\textbf{- w/o Refine} & 0.620 & 0.545 & 0.659 & 0.517 & \textbf{0.390} & 0.575 & 0.551 \\
\textbf{- w/o Multi-Step} & \underline{0.629} & 0.511 & 0.663 & 0.515 & 0.366 & 0.443 & 0.521 \\
\textbf{- w/o Adaptation} & 0.611 & 0.517 & 0.619 & 0.485 & 0.375 & \underline{0.579} & 0.531 \\

\bottomrule
\end{tabular}
\caption{\label{tab:ab_adaptation}Performance (\%) of CrossICL based on Llama3.1-8B with various demonstration adaptation.}
\end{table*}

\subsection{The Impact of Different Minimum Gap Selection Criteria}
\label{sec:diff_select}

In addition to the task description, input, and output, we further identify other demonstration selection criteria in previous ICL studies \citep{min-etal-2022-rethinking, an-etal-2023-context}, including demonstration diversity, structural similarity, and complexity similarity.
We evaluate the impact of all these criteria to further explore how to select source task demonstrations.
Specifically, the following criteria are discussed:
1) \textbf{\emph{taskdes}}: similarity of the task description in the query;
2) \textbf{\emph{taskinput}}: similarity of the task input in the query;
3) \textbf{\emph{output}}: similarity of the output;
4)~\textbf{\emph{diversity}}: diversity of the query;
5) \textbf{\emph{length}}: similarity of the query length;
6) \textbf{\emph{template}}: similarity of the task template. \textbf{\emph{length}} and \textbf{\emph{template}} correspond to structural similarity;
7) \textbf{\emph{complexity}}: similarity of the text complexity of the query.
Among them, \textbf{\emph{template}} and \textbf{\emph{taskdes}} are task-level criteria, while the others are example-level criteria.
We replace our demonstration selection (\S\ref{sec:memory}) with these criteria or their combinations, and the implementation details are provided in Appendix~\ref{diffselectimplet}.
Besides, \textbf{\emph{random}} refers to randomly selecting demonstrations.
Figure~\ref{fig:diff_select_our} shows the performance of CrossICL with 13 different selection criteria.

It can be observed that, on average, the \textbf{\emph{taskdes+taskinput}} (i.e., the original setting of our CrossICL) performs the best. This is because this setup reduces the gap between source and target queries, making subsequent alignment steps (\S\ref{sec:adaptation}) easier and applicable to a variety of task types.

Besides, different categories exhibit distinct preferences. Extraction and rewriting are sensitive to the query length because they typically involve paragraph-level inputs. If the demonstration is at the sentence level, it will provide an incorrect input distribution. Similarly, dialogue and generation prefer the similarity of outputs because these tasks often produce long and complex outputs.

Moreover, adding other criteria such as \textbf{\emph{diversity}} to \textbf{\emph{taskdes+taskinput}} does not improve the average performance. This is primarily because it sacrifices the weights of the \textbf{\emph{taskinput}}, which may increase the gap between source and target queries.

\subsection{Ablation Study on Progressive Task Adaptation}
\label{sec:as-ctdr}
We analyze the effects of various modifications to the demonstration adaptation (\S\ref{sec:adaptation}) in our CrossICL:
1) \textbf{w/o Src in LG}: Replace source-task guided label generation with zero-shot inference;
2)~\textbf{w/o Src in All}: Do not reference any source tasks, the LLMs generate new queries from scratch, refine them, and generate their labels;
3) \textbf{w/o Refine}: Remove the query refinement;.
4) \textbf{w/o Multi-Step}: Replace the original adaptation stage with a one-step Prompt-\ref{pt:p6} at Appendix~\ref{appendix:prompts};
5) \textbf{w/o Adaptation}: Remove the adaptation stage and directly use the selected source task demonstrations for ICL. As shown in Table~\ref{tab:ab_adaptation}, it can be found that:

First, when we remove the source task, the performance of CrossICL degrades. This is primarily because removing the source task leads to a lack of guidance in demonstration acquisition, reducing the quality of demonstrations.

Second, query refinement is helpful. Queries from the initial transforming step often contain noise, which requires further refinement to alleviate. Similarly, performance degradation from single-step adaptation occurs because the generated queries contain noise and are not refined, undermining label generation quality.

Additionally, directly using source task demonstrations for ICL hurts model performance. We also test with 7 other demonstration selection methods in Appendix~\ref{app:diff_select_notransfer} and arrive at the same conclusion. A further discuss can be found in \S\ref{sec:crosstaskerroranalyze}.

\subsection{Error Analysis of the Cross-Task Demonstration Interference}
\label{sec:crosstaskerroranalyze}

As shown in \S\ref{sec:as-ctdr} and Appendix \S\ref{app:diff_select_notransfer}, directly utilizing source task samples as target task ICL demonstrations harms the performance. Here, we analyze the types of interference caused by them.

For each task category, we examine 200 error cases, and the main types of interference we identified are as follows (examples are presented in Appendix~\ref{appendix:errorexample}):
1) \textbf{Misleading Task Type}: The model responds to the target task query directly based on the source task description;
2) \textbf{Confusing Output Format}: When the output formats of the target and source tasks are similar, the model may replicate the output format of source tasks, including label space, output constraints (e.g., capitalization), imitation of text style and length, etc; 3) \textbf{Keyword Shift}: Same keywords in both tasks may have different meanings. For example, both tasks involve determining whether an input is ``acceptable'', one might refer to the correctness of an answer, while the other is the answerability of a question. The model may mistakenly interpret the keyword based on the source task; 4) \textbf{Label Distribution Shift}: When source and target task output labels share similar semantics, the label distribution of the former may bias the latter. For example, the source task has few ``positive'' labels while the target task leans toward ``True'' labels; 5) \textbf{Simplified Thinking}: If the source task is simple, LLMs may approach the complex target task in an overly direct manner. We examine and find that some cases significantly shorten the CoT chain length. Besides, while there may be no obvious traces, the two potential interferences are: 6) \textbf{Incompatible Thinking Pattern}: Source task demonstrations may lead the LLMs to focus on information that is helpful for the source task but not for the target task, forming an inappropriate attention pattern; 7) \textbf{Information Overload}: The redundant tokens of source tasks may  interfere with the attention of LLMs.

\begin{figure}[t]
  \centering
\includegraphics[width=1\linewidth,trim=0 0cm 0 0cm, clip]{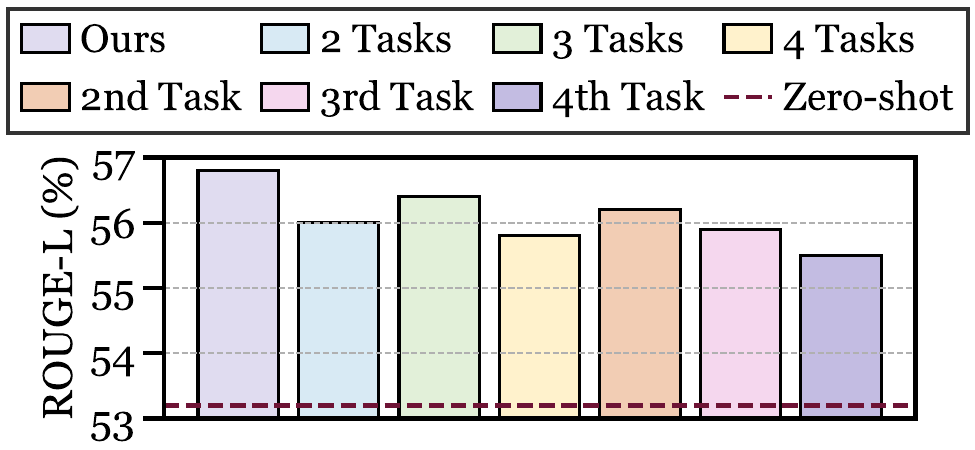}
  \caption{Average Performance (\%) of our CrossICL based on Llama3.1-8B with different source tasks.}
  \label{fig:diff_src}
\end{figure}

\subsection{The Impact of Source Task Similarity}
As shown in Figure~\ref{fig:diff_src}, we analyze the impact of selecting different source tasks in \S\ref{sec:memory} on our CrossICL: 1) \textbf{N Tasks} means that the number of selected tasks increases from 1 to $N$; 2) \textbf{$K$-th Task} means selecting the $K$-th ranked task instead of the first task.
It can be observed that \textbf{even when using source tasks that are less similar to the target task, CrossICL still effectively improves the model performance}. This is primarily due to our two-stage alignment-enhanced tolerance for the gap between source and target tasks.

\subsection{The Impact of the Number of Source Task Demonstrations}

\begin{figure}[t]
  \centering
\includegraphics[width=1\linewidth,trim=0 0.4cm 0 0.5cm, clip]{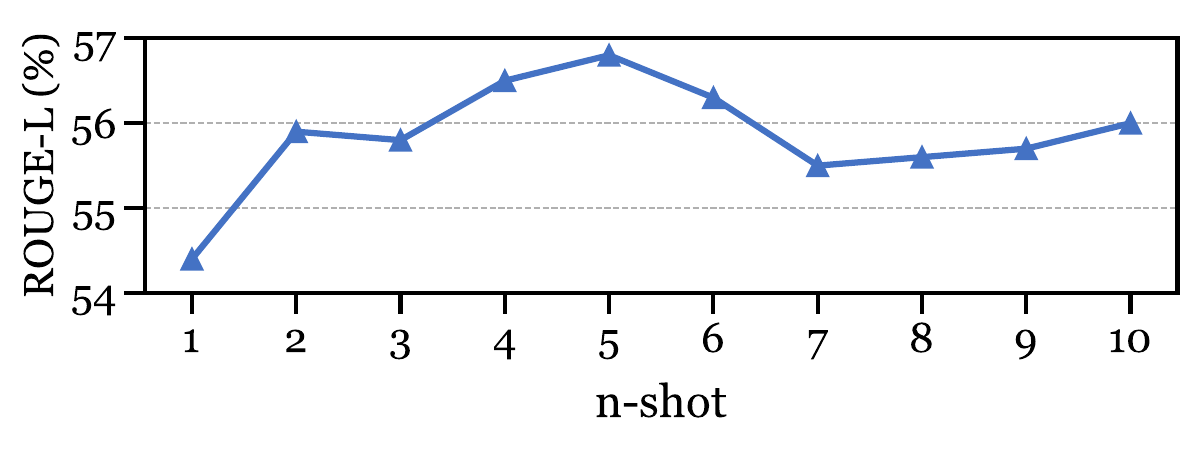}
  \caption{Average Performance (\%) of our CrossICL based on Llama3.1-8B with different amounts of aligned source task demonstrations.}
  \label{fig:nshot}
\end{figure}

As shown in Figure~\ref{fig:nshot}, we analyze the impact of using different numbers of source task demonstrations on CrossICL.
It can be observed that, from 1-shot to 10-shot, the performance of our CrossICL generally increases first and then decreases. This is mainly because as the number of demonstrations increases, the model receives more guidance. However, when there are too many demonstrations, they may hinder the model from focusing on the target query, interfering with its reasoning.

\begin{figure}[t]
  \centering
\includegraphics[width=1\linewidth,trim=0.3cm 0.1cm 0.5cm 0.2cm, clip]{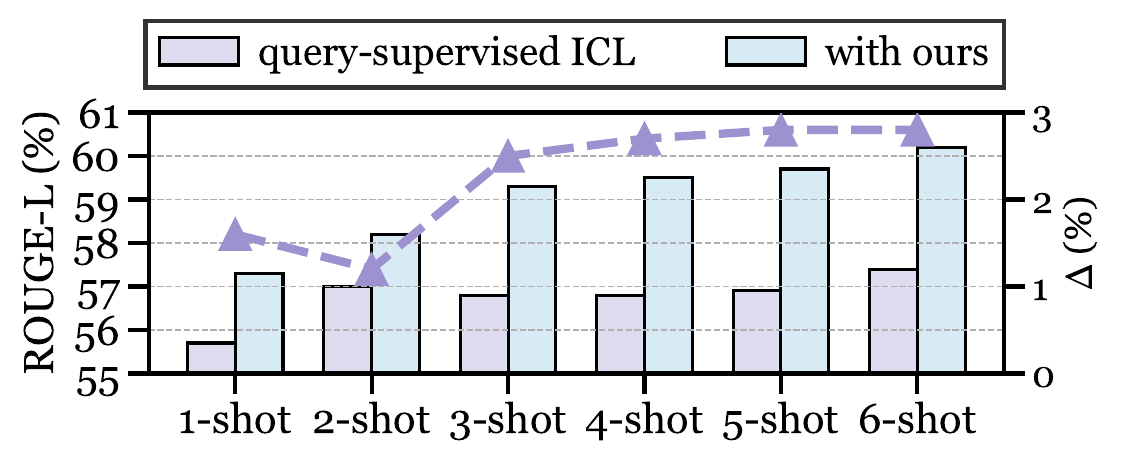}
  \caption{Average Performance (\%) of query-supervised ICL with our CrossICL based on Llama3.1-8B.}
  \label{fig:dail}
\end{figure}

\subsection{Combine with Query-Supervised ICL}
\label{sec:qsicl}

\citet{su-etal-2024-demonstration} highlight a special zero-shot ICL, where all user queries belong to the same task. They combine these queries with the labels generated by LLMs in zero-shot inference, which are then used as demonstrations for ICL to answer the query. We refer to this as query-supervised ICL. As shown in Figure~\ref{fig:dail}, on Super-NI, we combine this approach with CrossICL, i.e., using CrossICL generates labels for each query. It can be observed that CrossICL enhances the performance of query-supervised ICL. This is mainly because CrossICL, by learning from source tasks, generates more accurate labels for the constructed demonstrations.

\section{Related Work}

\subsection{Zero-shot ICL}
\label{related:zsicl}

Recent zero-shot ICL studies guide LLMs in automatically acquiring demonstrations.

SG-ICL \citep{kim2022self} guides LLMs to fill a predefined set of question templates and labels, which necessitates additional manual effort.
Z-ICL \citep{lyu2023z} selects sentences from external corpora as new questions and assigns random labels to obtain demonstrations with noise.
And, both SG-ICL and Z-ICL can only be applied to a subset of classification tasks.
Besides, Self-ICL \citep{chen2023self} guides LLMs in generating demonstrations entirely based on the task description, enabling its use across a wide range of tasks.
However, the absence of explicit guidance can compromise the quality of its generated demonstrations, leading to diminished model performance \citep{su-etal-2024-demonstration}.

In contrast, our CrossICL not only eliminates the need for additional manual effort and uses higher-quality demonstrations, but also extends in-task analogy of standard ICL to cross-task transfer. This broadens the application of ICL and offers a new pathway for generalization of LLMs. Besides, it is applicable to various NLP tasks.

\subsection{Semi-supervised ICL}
Beyond zero-shot ICL, some semi-supervised ICL studies aim to reduce rather than eliminate the cost of obtaining ICL demonstrations.

\citet{zhang2022automatic} and \citet{wang2024context} focus on the CoT demonstrations.
\citet{zhang2022automatic} utilize zero-shot CoT \citep{zscot} to generate reasoning chains for manually annotated examples.
\citet{wang2024context} utilize manually annotated CoT examples as references to guide LLMs in generating more CoT examples with data retrieved from external corpora.
Besides, \citet{su-etal-2024-demonstration} assume that users only input a single type of task. They utilize user past queries and their LLMs responses as demonstrations for subsequent queries.

Moreover, \citet{chatterjee-etal-2024-language} also discuss ICL in a cross-task manner but requires much user effort, degrading user satisfaction with ICL services. For a user query, they require users to manually annotate multiple similar new queries. Then, they employ existing demonstrations of all other task for ICL to repeatedly predict labels of these new queries, with weighted voting to determine the final label. Finally, these query-label pairs are used as demonstrations for the initial query. This method is only applicable to classification tasks because of its weighted voting strategy. In contrast, our work does not restrict task types and performs ICL under an \emph{unsupervised} cross-task transfer learning setting, rather than requiring much more user effort.

The above methods all rely on manually providing demonstrations for each target task. Considering the growing types of user tasks for LLMs today, this approach is not feasible enough. In contrast, CrossICL eliminates the need for additional manual effort to obtain the target task demonstrations, expanding the application potential of ICL.

\section{Conclusion}

In this work, we propose a new ICL paradigm CrossICL with unsupervised cross-task transferred existing demonstrations.
This eliminates the need for additional manual effort, providing a new avenue for the generalization of LLMs and expanding the application prospects of ICL.
To study CrossICL, we first design a concise and universal implementation as the foundation for experimental exploration. Then, we conduct a comprehensive experimental analysis of CrossICL, providing a wealth of valuable insights.


\section*{Limitations}
\label{sec:limit}

The Query Format is Limited to Two Parts. As stated in \S\ref{sec:prel}, following previous zero-shot ICL study \citep{chen2023self}, we require that the user query consists of two parts: the task description and the task input. This does indeed impose certain constraints on the format of user queries.
However, \textbf{the additional action introduced is minimal}, as users only need to distinguish these two parts when designing prompts, or LLM service providers can directly provide separate input fields for the two parts in the user interface.

Our method does \textbf{NOT} expect users to provide a stricter or more complete task description than what they typically use in their prompts.
We simply encourage users to place the main part describing the task objective at the beginning. In practical applications, this is very easy for users to achieve.

Besides, LLMs can also be utilized to automatically transform a user query into the two parts. For instance, \citet{gao-etal-2024-self-evolving} demonstrate that a simple prompt can reliably guide LLMs to generate task descriptions for a given query.

Moreover, our approach primarily performs cross-task transfer at the example level, while recent work \citep{gao2024selfevolvinggptlifelongautonomous,gao2025expetransllmsexperientialtransfer} has also explored cross-task transfer at the level of concepts or abstract task execution skills. The combination of both may further enhance the cross-task generalization capability of LLMs.

\bibliography{custom}

\clearpage

\appendix

\section*{Appendix}

\begin{figure*}[t]
  \centering
\includegraphics[width=1\linewidth,trim=0 0cm 0 0cm, clip]{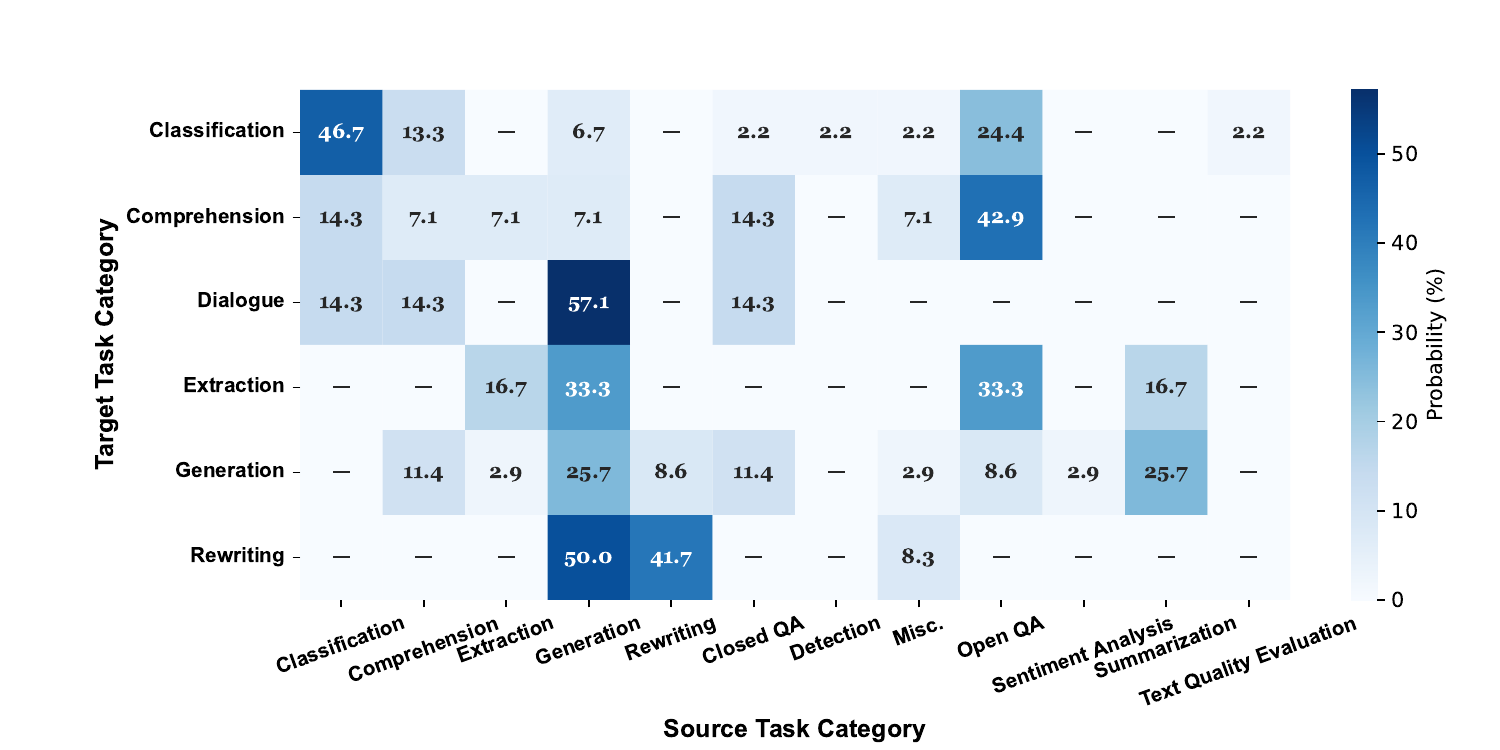}
  \caption{The distribution (\%) of source-target task pairs. Each value represents the probability of selecting the current source task type based on the current target task type.}
  \label{fig:hotmap}
\end{figure*}

\begin{table*}[!t]
\small
\centering
\setlength{\tabcolsep}{6pt}
\begin{tabular}{lccccccc}
\toprule
\textbf{Method} & \textbf{Classification} & \textbf{Comprehension} & \textbf{Dialogue} & \textbf{Extraction} &  \textbf{Generation} & \textbf{Rewriting} & \textit{\textbf{Avg.}} \\
\midrule

  \textbf{Ours} & \textbf{0.635} & \textbf{0.566} & \textbf{0.693} & \textbf{0.539} & \textbf{0.383} & \textbf{0.593} & \textbf{0.568} \\
 \textbf{- w/ 2 Src Demos} & 0.621 & 0.534 & 0.664 & 0.520 & \underline{0.381} & 0.587 & 0.551 \\
 \textbf{- w/ 3 Src Demos} & \underline{0.624} & \underline{0.538} & \underline{0.679} & \underline{0.521} & 0.381 & \underline{0.591} & \underline{0.556} \\
\bottomrule
\end{tabular}
\caption{\label{tab:sglg}Performance (\%) of our CrossICL based on Llama3.1-8B with multi-shot Source-Guided Label Generation.}
\end{table*}

\section{Analysis of Source-Guided Label Generation}
As shown in Table~\ref{tab:sglg}, we analyze whether using more source task demonstrations is helpful in the Source-Guided Label Generation step. It can be observed that introducing more source task demonstrations actually harms the performance of our CrossICL. This is because the additional demonstrations introduced do not have a close relationship with the aligned query, and only the corresponding source task demonstration before alignment are more likely to provide reliable guidance for label generation of the aligned query.

\section{The Distribution of the Source-Target Task Pairs}
As shown in Figure~\ref{fig:hotmap}, we show the distribution of different source-target task pairs in our demonstration selection stage. Specifically, each value represents the probability of selecting the corresponding source task based on the current target task. ``-'' indicates that the combination does not appear in our experiments. We follow the task categorization of \citet{wang-etal-2024-inscl}.

It can be observed that combinations of tasks within the same type occur most frequently. Moreover, transfer often occurs between task types that are evidently similar. This indicates that our implementation effectively identifies closely related tasks, thereby reducing the differences between the source and target tasks.

\section{Demonstration-Selection-Based ICL Methods}
\label{app:diff_select_notransfer}

\begin{figure}[t]
  \centering
\includegraphics[width=1\linewidth,trim=0 0.5cm 2.5cm 0.5cm, clip]{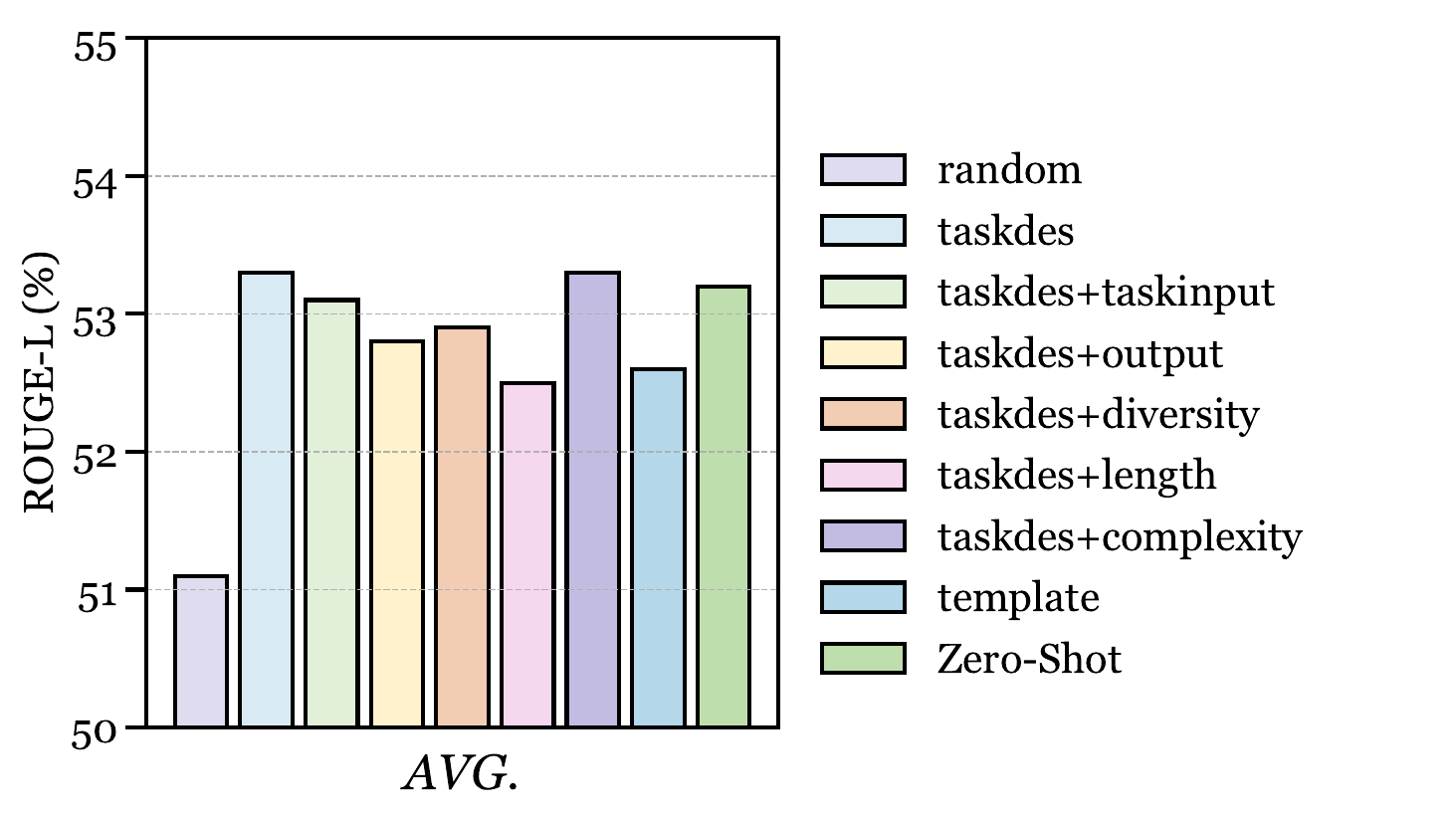}
  \caption{Performance (\%) of Cross-Task ICL based on Llama3.1-8b with unsupervised selected cross-task demonstrations.}
  \label{fig:notransfer_diff_select}
\end{figure}

\paragraph{Selection with Unsupervised Criteria.}
As illustrated in Figure~\ref{fig:notransfer_diff_select}, we evaluate the ICL directly using the selected source task demonstrations without any additional adaptation (\S\ref{sec:adaptation}), including various selection settings (consistent with those in \S\ref{sec:diff_select} and Appendix~\ref{diffselectimplet}). ``Zero-Shot'' refers to the results of the LLMs performing zero-shot inference. It can be observed that, although there are differences across selection settings, except for random selection, the performance of other settings is similar, and none can effectively surpass the zero-shot baseline method.

\paragraph{Selection Based on Supervised ICL Retriever.}
Based on previous studies, supervised ICL retrievers tend to perform better than unsupervised ICL selection \citep{luo2024context}. Therefore, we further discuss the supervised ICL retriever-based selection methods under the cross-task setting, taking LLM-R \citep{wang2024learning} as a representative. LLM-R is one of recent most powerful ICL retriever, specifically trained for demonstration selection based on a wide range of task datasets.

As shown in Table~\ref{tab:app_selection_baselines}, \textbf{LLM-R} refers to directly using LLM-R for selection and applying the selected demonstrations for ICL; while \textbf{Ours w/ LLM-R} refers to replacing the selection module in our CrossICL with LLM-R, that is, further adapting the LLM-R-selected demonstrations via our Progressive Task Adaptation.

It can be observed that the performance of demonstration selection based on LLM-R is suboptimal, possibly because LLM-R is not specifically designed for cross-task scenarios and is trained on intra-task data, which harms its performance in the cross-task setting.
To the best of our knowledge, there is currently no supervised ICL retriever specifically designed for cross-task scenarios.
Moreover, it is also evident that directly using the selected source-task demonstrations for ICL without our proper adaptation does not achieve optimal performance. The possible reasons are discussed in \S\ref{sec:crosstaskerroranalyze}.

\begin{table*}[!t]
\small
\centering
\setlength{\tabcolsep}{6pt}
\begin{tabular}{lccccccc}
\toprule
\textbf{Method} & \textbf{Classification} & \textbf{Comprehension} & \textbf{Dialogue} & \textbf{Extraction} &  \textbf{Generation} & \textbf{Rewriting} & \textit{\textbf{Avg.}} \\
\midrule
\textbf{Zero-Shot} & 0.602 & 0.512 & 0.655 & 0.507 & \textbf{0.398} & 0.520 & 0.532 \\
\textbf{LLM-R} & 0.606 & 0.473 & 0.656 & 0.501 & 0.383 & 0.534 & 0.526 \\
\textbf{Ours w/ LLM-R} & \textbf{0.643} & \underline{0.532} & \underline{0.674} & \textbf{0.542} & \underline{0.386} & \underline{0.544} & \underline{0.553} \\
\textbf{Ours} & \underline{0.635} & \textbf{0.566} & \textbf{0.693} & \underline{0.539} & 0.383 & \textbf{0.593} & \textbf{0.568} \\

\bottomrule
\end{tabular}
\caption{\label{tab:app_selection_baselines}Performance (\%) of Cross-Task ICL based on Llama3.1-8B with supervised ICL retriever.}
\end{table*}

\begin{table}[!t]
\small
\centering
\setlength{\tabcolsep}{5pt}
\begin{tabular}{lcccc}
\toprule
\multirow{2}{*}{\textbf{Dataset}} & \multicolumn{2}{c}{\textbf{Annotation Time (s)}}& \multicolumn{2}{c}{\textbf{Performance (\%)}}\\
\cmidrule(lr){2-3} \cmidrule(lr){4-5}
& \textbf{Manual} & \textbf{Ours}& \textbf{Manual} & \textbf{Ours}\\
\midrule
 \textbf{MC-TACO} & 94 & 6.6 & 0.580 & 0.591 \\
\textbf{COPA} & 125 & 7.2 & 0.890 & 0.931 \\
\textbf{TweetQA} & 62 & 8.6 & 0.950 & 0.970 \\
\textbf{e-SNLI} & 83 & 7.5 & 0.540 & 0.570 \\
\bottomrule
\end{tabular}
\caption{\label{tab:costsvshuman}Annotation time (s) and performance (\%) of the ICL with the demonstrations obtained by the manual annotation and our CrossICL based on Llama3.1-8B.}
\end{table}

\section{Discussion on the Efficiency with Standard ICL Settings}
\label{app:costcomp}
Taking several example tasks in Super-NI as cases, we discuss the cost issues currently faced by ICL and the improvements introduced by our CrossICL. The tasks involved are as follows: 1) MC-TACO \citep{zhou2019going}, which aims to test the reasoning ability for temporal commonsense, consisting of five time-related question categories (duration, temporal order, typical time, frequency, and staticity); 2) COPA \citep{roemmele2011choice}, used to assess causal reasoning ability, requiring the model to choose the more reasonable cause or effect from two similar options given a premise; 3) TweetQA \citep{xiong2019tweetqa}, the question-answering dataset focusing on social media (particularly Twitter); 4) e-SNLI \citep{NIPS2018_8163}, a classic NLI dataset used to test the textual reasoning ability.

Table~\ref{tab:costsvshuman} shows the average time required to manually annotate a single demonstration for these tasks, as well as the average time needed by our CrossICL to generate 5 demonstrations for a single query. Additionally, the table presents the performance of ICL using the manually annotated demonstration or those generated by our CrossICL. Note that, since the papers of these datasets do not describe the time required for annotation (in fact, most dataset papers do not report time costs), for each dataset, we manually annotate 10 examples following their annotation processes and record the average time, providing an approximate estimate of the time cost. Besides, we do not include the time required to prepare the data for annotation, such as collecting relevant news corpora, tweet corpora, etc. This often requires a significant amount of time and demands that users have certain programming skills and data access permissions, which further hinders the application of ICL. It can be observed that even without considering the obvious time cost of preparing data for manual annotation, our approach still effectively alleviates the time cost of manual annotation and achieves comparable performance.

\begin{figure}[t]
  \centering
\includegraphics[width=1\linewidth,trim=1.2cm 1.2cm 0 1cm, clip]{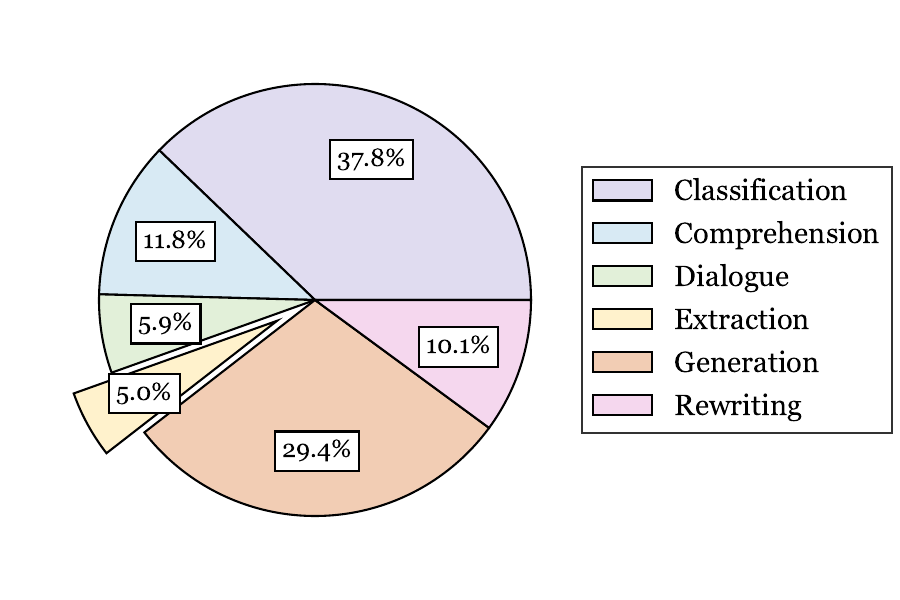}
  \caption{The distribution (\%) of the Super-NI test set.}
  \label{fig:supernidistrib}
\end{figure}

\section{Task Category Distribution}
\label{superni-more}
As shown in Figure~\ref{fig:supernidistrib}, we report the task category distribution in the Super-NI test set. Besides, following \citet{wang-etal-2024-inscl}, the training set consists of 16 major categories: Classification 16.9\%, Generation 15.7\%, Program Execution 11.9\%, Open QA 11.5\%, Closed QA 10.6\%, Detection 7.1\%, Sentiment Analysis 5.6\%, Misc. 4.8\%, Extraction 4.1\%, Comprehension 3.4\%, Rewriting 2.6\%, Code 2.1\%, Summarization 1.6\%, Text Quality Evaluation 0.9\%, Dialogue 0.5\%, Mathematics 0.5\%.
This dataset includes a wide variety of task types, such as causal reasoning, which have been verified to remain challenging for current LLMs to accomplish \citep{gao2023chatgptgoodcausalreasoner,10.1007/978-981-97-9440-9_6,GAO2024112139}.


\section{Implementation of Different Demonstration Selection Settings}
\label{diffselectimplet}

Here, we describe the specific implementation\footnote{For the convenience of readers, we describe each setting from the beginning, which may result in some repetition texts across different settings.} of the different demonstration selection settings in \S\ref{sec:diff_select} and Appendix~\ref{app:diff_select_notransfer}.
All semantic similarities below are calculated using the cosine similarity of text embedding vectors.
1)~\textbf{\emph{random}}: Randomly select from all existing candidate source task demonstrations.
2)~\textbf{\emph{taskdes}}: Select the most similar source task based on the semantic similarity of the task description, and then randomly choose demonstrations from that task.
3)~\textbf{\emph{taskdes+taskinput}}: No modifications are made, i.e., the original selection strategy of our framework is utilized.
4)~\textbf{\emph{taskdes+output}}: Select the most similar source task based on the semantic similarity of the task description, and then, choose the demonstrations with the most similar label based on the semantic similarity between the golden label of the source task demonstration and the golden label of the target query.
5)~\textbf{\emph{taskdes+diversity}}: Select the most similar source task based on the semantic similarity of the task description, use the \emph{K-means} algorithm \citep{macqueen1967some} to cluster its demonstrations, and select one demonstration from each cluster to enhance diversity of the selected set of demonstrations.
6)~\textbf{\emph{taskdes+length}}: Select the most similar source task based on the semantic similarity of the task description, and then choose source task demonstrations with query lengths closest to the target task query. The lengths are measured in tokens.
7)~\textbf{\emph{taskdes+complexity}}: Select the most similar source task based on the semantic similarity of the task description, and then choose source task demonstrations whose text complexity of the query is closest to that of the target task query. We use the perplexity calculated by LLMs themselves to measure the text complexity. Perplexity reflects how complex or challenging a text is from the perspective of LLMs.
8)~\textbf{\emph{template}}: We first use Prompt-\ref{pt:p5} to guide LLMs in summarizing the template for each task. Then, we select the most similar source task based on the semantic similarity of the task template, and randomly choose demonstrations from that task.
9)~\textbf{\emph{taskdes+taskinput+output}}: We first select the most similar source task based on the semantic similarity of the task description. Then, we rank the source task demonstrations based on their semantic similarity to the task input of the target task query and the semantic similarity of their golden labels. The harmonic mean of these two rankings is calculated. Finally, we select the highest-ranked source task demonstrations from the averaged ranking.
10)~\textbf{\emph{taskdes+taskinput+diversity}}: We first select the most similar source task based on the semantic similarity of the task description. Then, we select the top 100 source task demonstrations with task inputs that are most semantically similar to that of the target task query. Afterward, we use K-means clustering on these 100 demonstrations and select one from each cluster to enhance the diversity of the selected set.
11)~\textbf{\emph{taskdes+taskinput+length}}: We first select the most similar source task based on the semantic similarity of the task description.
Next, the source task demonstrations are ranked based on two aspects: the semantic similarity between the task inputs of the source task demonstrations and the target task query, and the similarity in the lengths of their queries.
Then, the harmonic mean of these two rankings is calculated.
Finally, the highest-ranked source task demonstrations based on the average ranking is selected.
12)~\textbf{\emph{taskdes+taskinput+complexity}}: We first select the most similar source task based on the semantic similarity of the task description. Then, we rank the source task demonstrations according to the semantic similarity between the task inputs of the source task demonstrations and the target task query, as well as the similarity in query complexity.
The harmonic mean of these two rankings is then calculated. Finally, the highest-ranked source task demonstrations based on the average ranking are selected.
13)~\textbf{\emph{template+taskinput}}: We first use Prompt-\ref{pt:p5} to guide LLMs in summarizing the template for each task. Then, we select the most similar source task based on the semantic similarity of the task template, and choose source task demonstrations whose task inputs are most semantically similar to the target task query.

\begin{table*}[!t]
\small
\centering
\setlength{\tabcolsep}{6pt}
\begin{tabular}{llcccc}
\toprule
\textbf{Model} & \textbf{Method} & \textbf{Classification} & \textbf{Comprehension} & \textbf{Dialogue} & \textbf{Extraction} \\
\midrule
\multirow{4}{*}{\textbf{Llama3.1-8B}}
& \textbf{Zero-Shot} & 0.538 & 0.413 & 0.619 & 0.227 \\
& \textbf{Zero-shot-CoT} & \textbf{0.586} & \underline{0.442} & \underline{0.651} & 0.190 \\
& \textbf{Self-ICL} & 0.544 & 0.432 & 0.614 & \underline{0.230} \\
& \textbf{Ours} & \underline{0.581} & \textbf{0.487} & \textbf{0.661} & \textbf{0.255} \\
\midrule
\multirow{4}{*}{\textbf{Gemma2-9B}}
 & \textbf{Zero-Shot} & 0.553 & \underline{0.480} & 0.596 & \textbf{0.285}\\
& \textbf{Zero-shot-CoT} & 0.556 & \textbf{0.523} & 0.567 & 0.235 \\
& \textbf{Self-ICL} & \underline{0.559} & 0.478 & \underline{0.613} & 0.278  \\
& \textbf{Ours} & \textbf{0.618} & 0.469 & \textbf{0.696} & \underline{0.280}  \\
\midrule
\multirow{4}{*}{\textbf{Qwen2-7B}}
 & \textbf{Zero-Shot} & 0.642 & 0.453 & 0.650 & 0.228 \\
& \textbf{Zero-shot-CoT} & 0.553 & \underline{0.460} & 0.596 & 0.203\\
& \textbf{Self-ICL} & \textbf{0.651} & 0.446 & \underline{0.656} & \underline{0.235} \\
& \textbf{Ours} & \underline{0.648} & \textbf{0.481} & \textbf{0.667} & \textbf{0.250}\\
\midrule
\multirow{4}{*}{\textbf{Qwen2.5-7B}}
 & \textbf{Zero-Shot} & 0.507 & 0.429 & \underline{0.707} & 0.280  \\
& \textbf{Zero-shot-CoT} & \underline{0.580} & \textbf{0.510} & 0.700 & 0.265  \\
& \textbf{Self-ICL} & 0.532 & 0.424 & 0.700 & \underline{0.285} \\
& \textbf{Ours} & \textbf{0.653} & \underline{0.461} & \textbf{0.716} & \textbf{0.307} \\
\midrule
\multirow{4}{*}{\textbf{Deepseek-7B}}
 & \textbf{Zero-Shot} & 0.463 & 0.282 & \underline{0.479} & \underline{0.223}\\
& \textbf{Zero-shot-CoT} & 0.455 & \textbf{0.344} & 0.469 & 0.157  \\
& \textbf{Self-ICL} & \underline{0.472} & 0.299 & 0.466 & \textbf{0.232}  \\
& \textbf{Ours} & \textbf{0.474} & \underline{0.300} & \textbf{0.491} & 0.207  \\
\midrule
\multirow{4}{*}{\textbf{GPT-4o}}
 & \textbf{Zero-Shot} & 0.692 & 0.617 & 0.753 & 0.252 \\
& \textbf{Zero-shot-CoT} & \underline{0.716} & \underline{0.635} & 0.750 & 0.250  \\
& \textbf{Self-ICL} & 0.710 & 0.619 & \underline{0.755} & \textbf{0.270}  \\
& \textbf{Ours} & \textbf{0.739} & \textbf{0.673} & \textbf{0.772} & \underline{0.253}\\
\bottomrule
\end{tabular}
\caption{\label{tab:main-acc}\textbf{Accuracy (exact match, \%)} on the Super-NI benchmark. \textbf{Bold} and \underline{Underlined} represent the 1st and the 2nd best-performing methods for each LLM. Please note that we report the exact match accuracy metric here, \textbf{which is not applicable to generation and rewriting tasks} (these two tasks are included in Table~\ref{tab:main}, where ROUGE-L is used for evaluation).}
\end{table*}

\begin{figure}[t]
  \centering
\includegraphics[width=1\linewidth,trim=2cm 1cm 0cm 1cm, clip]{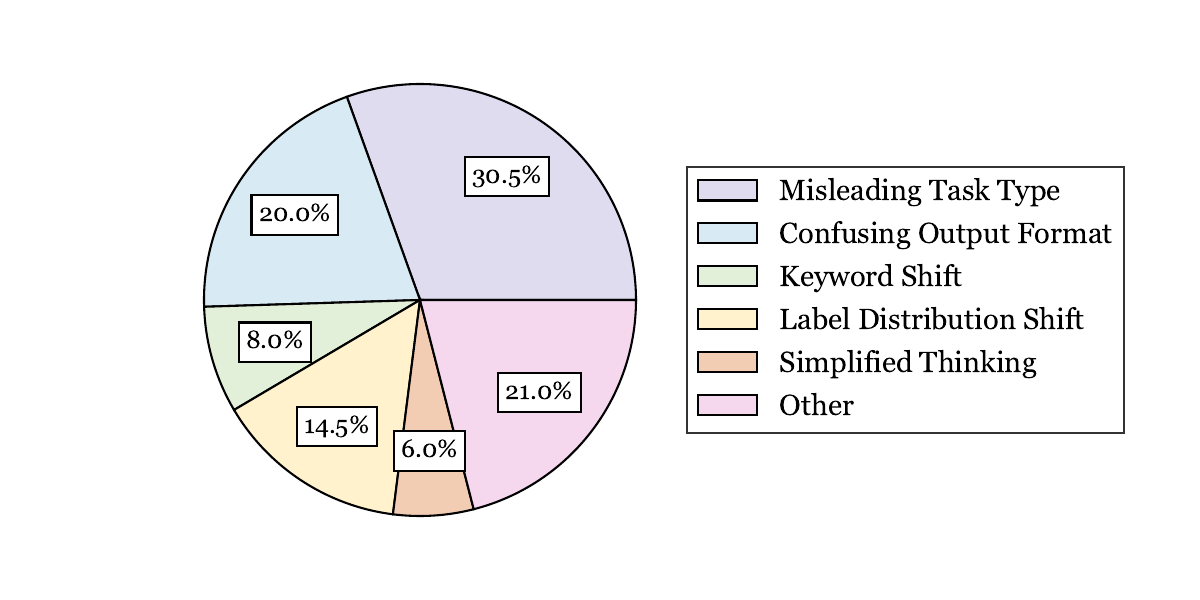}
  \caption{The error distribution (\%) of the types of cross-task demonstration interference. ``Other'' refers to the proportion of cases where the reasons for the errors are unclear.}
  \label{fig:error-distrib}
\end{figure}

\section{Performance with Exact Match Accuracy}
\label{app:em}
In the previous sections, we use ROUGE-L as a unified evaluation metric across all tasks. A question is: \textbf{whether ROUGE-L is a reliable metric for tasks such as classification in Super-NI?}

First, \textbf{the Super-NI dataset is specifically designed by its authors with the ROUGE-L metric in mind.} It provides multiple possible answers for each question to ensure that ROUGE-L can effectively evaluate all tasks included in the dataset. This point is explicitly emphasized in the original paper or Super-NI, which even includes a dedicated subsection to validate the reliability of the ROUGE-L metric on Super-NI.

Second, in this section, \textbf{we report the performance using the Exact Match (i.e., Accuracy) metric on the Super-NI dataset to further validate the reliability of ROUGE-L for deterministic tasks in Super-NI.} Please note that we do not include generation and rewriting tasks, as they are inherently unsuitable for exact match but are appropriate for ROUGE-L.
Table~\ref{tab:main-acc} presents the experimental results.

By carefully comparing the results in Table~\ref{tab:main} for ROUGE-L and Table~\ref{tab:main-acc} for exact match Accuracy, we find that \textbf{the performance trends and relative rankings shown by ROUGE-L are highly consistent with those of the Accuracy metric}. This aligns with the experimental findings reported by the authors of the Super-NI.

\section{The Impact of Sampling Temperature}

\begin{figure}[b]
  \centering
\includegraphics[width=1\linewidth,trim=0 0cm 0cm 0cm, clip]{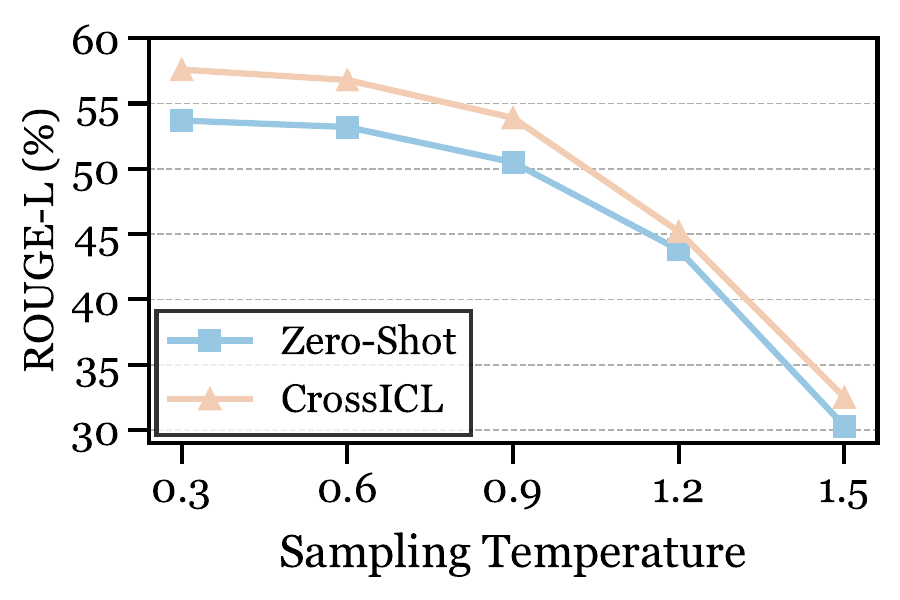}
  \caption{Performance (\%) of Cross-Task ICL based on Llama3.1-8b with different sampling temperatures.}
  \label{fig:diff_sample_T}
\end{figure}

As shown in the Figure~\ref{fig:diff_sample_T}, we analyze the effect of different sampling temperatures on ICL in the cross-task setting. Specifically, we report the average scores of LLMs on the Super-NI dataset under various temperatures.

It can be observed that, whether in zero-shot inference or in our CrossICL method, the performance of LLMs consistently declines as the temperature increases. This may be because higher sampling temperatures lead to increased randomness in the reasoning processes of LLMs, causing instability in the inference and reducing performance.
Besides, we also find that higher temperatures increase the runtime, as higher temperatures tend to produce more diverse and longer responses.

\section{Information on Responsible NLP Research}
\label{appendix:checklist}

\paragraph{Use Scientific Artifacts.}
In this work, we conduct experiments using the Super-NI benchmark dataset, as described in Section \S\ref{datasets}. We perform experiments only on the English data. The Super-NI dataset is allowed for scientific research, and its use aligns with its intended purpose.
The Super-NI benchmark follows the Apache License 2.0, and its demographic distribution does not exhibit significant bias. Besides, Super-NI has been widely used in the NLP community and has undergone review. It does not contain names or unique information that could identify individuals, nor does it include offensive content. Additionally, we randomly selected 5 samples from each test sub-task data for manual inspection to ensure that the datasets do not contain inappropriate content.

\paragraph{Potential Risks.} Our work discusses cross-task ICL, extending ICL from traditional task-specific analogies to task transfer, which enhances the capabilities of LLMs. Although there are no obvious risks, our approach could be used for malicious purposes, such as applying cross-task ICL in illegal online activities.

\paragraph{Computational Details}
As shown in \S\ref{sec:paramsetting}, we conduct experiments using six different LLMs, most of whose names already reflect the model size. The only exception is GPT-4o, but its exact scale has not been disclosed. The current available information suggests that GPT-3.5 likely contains around 175 billion parameters, and GPT-4o is larger than it. For the local experiment, we used an A100 PCIe 40GB, which roughly takes 6 hours.

\paragraph{Parameters For Packages} The version of packages we used: FlagEmbedding=1.3.2, torch=2.4.0, transformers=4.46.2, vllm=0.6.3, openai=1.54.3, numpy=1.26.4.
More details are provided in our supplementary materials.

\paragraph{Human Participants}
Overall, our work involved minimal human participation, with two exceptions: 1) the error analysis in \S\ref{sec:crosstaskerroranalyze}; 2) the annotation speed tests in \S\ref{app:costcomp}. These are carried out by our authors, as they are not part of an employment relationship (i.e., no compensation is involved) and do not constitute formal data annotation work. This is simply one of the ways the authors contributed to the paper.
The annotation tests in \S\ref{app:costcomp} follow the guidelines of the target dataset, such as risk warnings. All authors are aware of the intended use of the above data. No involvement from an ethics review board is needed. The data processing does not introduce racial bias or gender discrimination. Given the anonymity requirements, the geographical distribution of the authors will be disclosed after the paper is accepted.

\paragraph{AI Assistants in Writing.}
We only use AI to assist with language expression. Specifically, we use ChatGPT to help check for grammar errors and provide suggestions for improving the clarity and style of the language.

\section{Prompts}
\label{appendix:prompts}
In this section, we provide all the prompts used in our framework and analysis experiments.

\captionsetup[table]{name=Prompt} 
\captionsetup{labelformat=empty} 
\setcounter{table}{0} 

\begin{table*}[hb] \small 
\centering
\begin{tabular}{p{\linewidth-1cm}}
\toprule
\textbf{Prompt 1: Rewrite the source task query into a new target task query based on the user target task query.}\\
\midrule\begin{lstlisting}[escapeinside={(*@}{@*)}]
(*@\textbf{USER: }@*)
<Source Task>
<Source Task Instruction>
(*@\sethlcolor{lightblue}\hl{[source task description]}@*)
</Source Task Instruction>
<Source Task Query>
(*@\sethlcolor{lightblue}\hl{[source task input]}@*)
</Source Task Query>
</Source Task>

<Target Task>
<Target Task Instruction>
(*@\sethlcolor{lightblue}\hl{[target task description]}@*)
</Target Task Instruction>
<Example of Target Task Query>
(*@\sethlcolor{lightblue}\hl{[target task input]}@*)
</Example of Target Task Query>
</Target Task>

Please rewrite the Source Task Query to synthesize a new Target Task Query.

Output as follows:
<Rewrote>
<Target Task Query>
...
</Target Task Query>
</Rewrote>
\end{lstlisting}\\\bottomrule
\end{tabular}
\caption{}
\label{pt:p1}
\end{table*}

\begin{table*}[hb] \small 
\centering
\begin{tabular}{p{\linewidth-1cm}}
\toprule
\textbf{Prompt 2: Refine the synthesized query based on the user target task query.}\\
\midrule\begin{lstlisting}[escapeinside={(*@}{@*)}]
(*@\textbf{USER: }@*)
<Synthesized Query>
(*@\sethlcolor{lightblue}\hl{[synthesized query]}@*)
</Synthesized Query>

<Target Task>
<Target Task Instruction>
(*@\sethlcolor{lightblue}\hl{[target task description]}@*)
</Target Task Instruction>
<Example of Target Task Query>
(*@\sethlcolor{lightblue}\hl{[target task input]}@*)
</Example of Target Task Query>
</Target Task>

For the target task, I have synthesized a new query following the example of target task query. Could you help me refine the synthesized query?
Return your refined query in the following format:
<Refined Query>
...
</Refined Query>
\end{lstlisting}\\\bottomrule
\end{tabular}
\caption{}
\label{pt:p2}
\end{table*}

\begin{table*}[hb] \small 
\centering
\begin{tabular}{p{\linewidth-1cm}}
\toprule
\textbf{Prompt 3: Generate the labels of the synthesized queries by referring to the original source tasks demonstration.}\\
\midrule\begin{lstlisting}[escapeinside={(*@}{@*)}]
(*@\textbf{USER: }@*)
Note: You may refer to the following source task demonstrations, which might help you solve the task.

Source Task Instruction:
(*@\sethlcolor{lightblue}\hl{[source task description]}@*)

(*@\sethlcolor{lightblue}\hl{[source task input]}@*)

The final answer is: (*@\sethlcolor{lightblue}\hl{[source task golden label]}@*)



Please generate a response to the following target task question:

Target Task Instruction:
(*@\sethlcolor{lightblue}\hl{[target task description]}@*)

(*@\sethlcolor{lightblue}\hl{[target task input]}@*)
\end{lstlisting}\\\bottomrule
\end{tabular}
\caption{}
\label{pt:p3}
\end{table*}

\begin{table*}[hb] \small 
\centering
\begin{tabular}{p{\linewidth-1cm}}
\toprule
\textbf{Prompt 4: Perform standard ICL using source task demonstrations processed through our two-stage alignment strategy.}\\
\midrule\begin{lstlisting}[escapeinside={(*@}{@*)}]
(*@\textbf{USER: }@*)
Task Instruction:
(*@\sethlcolor{lightblue}\hl{[target task description]}@*)

(*@\sethlcolor{lightblue}\hl{[aligned task input]}@*)

The final answer is: (*@\sethlcolor{lightblue}\hl{[aligned task label]}@*)

(*@\sethlcolor{lightblue}\hl{... other aligned demonstrations ...}@*)

Task Instruction:
(*@\sethlcolor{lightblue}\hl{[target task description]}@*)

(*@\sethlcolor{lightblue}\hl{[current user target task input]}@*)
\end{lstlisting}\\\bottomrule
\end{tabular}
\caption{}
\label{pt:p4}
\end{table*}

\begin{table*}[hb] \small 
\centering
\begin{tabular}{p{\linewidth-1cm}}
\toprule
\textbf{Prompt 5: Summarize the template of the task inputs.}\\
\midrule\begin{lstlisting}[escapeinside={(*@}{@*)}]
(*@\textbf{USER: }@*)
For the following task and three examples of its inputs, please summarize the template of its inputs.
Return as follows:
<Input Template> ... </Input Template>

Task Instruction: (*@\sethlcolor{lightblue}\hl{[task description]}@*)

Input Example1:
(*@\sethlcolor{lightblue}\hl{[task input example 1]}@*)

Input Example2:
(*@\sethlcolor{lightblue}\hl{[task input example 2]}@*)

Input Example3:
(*@\sethlcolor{lightblue}\hl{[task input example 3]}@*)
\end{lstlisting}\\\bottomrule
\end{tabular}
\caption{}
\label{pt:p5}
\end{table*}

\begin{table*}[hb] \small 
\centering
\begin{tabular}{p{\linewidth-1cm}}
\toprule
\textbf{Prompt 6: The ablation setup used in \S\ref{sec:as-ctdr}, where progressive task adaptation is completed in a single-step prompt.}\\
\midrule\begin{lstlisting}[escapeinside={(*@}{@*)}]
(*@\textbf{USER: }@*)
<Source Task>
<Source Task Instruction>
(*@\sethlcolor{lightblue}\hl{[source task description]}@*)
</Source Task Instruction>
<Source Task Query>
(*@\sethlcolor{lightblue}\hl{[source task input]}@*)
</Source Task Query>
<Source Task Answer>
(*@\sethlcolor{lightblue}\hl{[source task answer]}@*)
</Source Task Answer>
</Source Task>

<Target Task>
<Target Task Instruction>
(*@\sethlcolor{lightblue}\hl{[target task description]}@*)
</Target Task Instruction>
<Example of Target Task Query>
(*@\sethlcolor{lightblue}\hl{[target task input]}@*)
</Example of Target Task Query>
</Target Task>

Please rewrite the Source Task Query and Answer to synthesize a new pair of Target Task Query and Answer.

Output as follows:
<Rewrote>
<Target Task Query>
...
</Target Task Query>
<Target Task Answer>
...
</Target Task Answer>
</Rewrote>
\end{lstlisting}\\\bottomrule
\end{tabular}
\caption{}
\label{pt:p6}
\end{table*}

\clearpage
\section{Framework Examples}
\label{appendix:case_study_our}
This section presents execution examples of our framework. Here is a brief summary of the operation process to help understand the examples:

Overall, for each user query, we first perform Minimum Gap Selection, choosing the source task demonstrations with the smallest gap to the user query. This involves first calculating the semantic similarity of task descriptions to identify the most similar source task, and then from the selected task, selecting the top-K most similar source task examples based on the semantic similarity of the demonstrations.

Then, we conduct Progressive Task Adaptation, which involves multi-stage transformation to achieve further alignment. Specifically, we first use prompts to guide LLMs in rewriting the source task examples to match the form of the user's target task. Next, we employ prompts to instruct LLMs to optimize the rewritten examples and correct any formatting errors. Finally, we use prompts to guide LLMs in generating answers for the optimized examples by referencing the answers from the source task examples.

Finally, we employ the aligned demonstrations for ICL to answer the initial user query.

\captionsetup[table]{name=ExampleofPrompt} 
\captionsetup{labelformat=empty} 
\setcounter{table}{0} 

\begin{table*}[hb] \small 
\centering
\begin{tabular}{p{\linewidth-1cm}}
\toprule
\textbf{Example of Prompt 1: Rewrite the source task query into a new target task query based on the user target task query.}\\
\midrule\begin{lstlisting}[escapeinside={(*@}{@*)}]
(*@\textbf{USER: }@*)
<Source Task>
<Source Task Instruction>
You are given a question or fill-in-the-blank question, two answer options (Option1 and Option2) and an Explanation. Your task is to find the correct answer (return the string of the correct option, not option1/2) for the given question from the given options and using explanation.
</Source Task Instruction>
<Source Task Query>
Input:
Question: Dane was on a mountaintop and struggled to breathe, so he rectified it by getting to an elevation that was
 Option1: higher
 Option2: lower
 Explanation: At higher elevations, there is less air to press on a given area.

Give your final answer in the following format: "The final answer is: [your answer]"
</Source Task Query>
</Source Task>

<Target Task>
<Target Task Instruction>
Given a premise and two alternatives, choose the alternative that is a more plausible cause or effect of the situation described by the premise. The input format is "premise
(1)alternative_1(2)alternative_2", the output should either be "1" or "2" based on your judgment.
</Target Task Instruction>
<Example of Target Task Query>
Input:
I ran out of breath.
(1)I climbed several flights of stairs.(2)I read several chapters of the book.

Give your final answer in the following format: "The final answer is: [your answer]"
</Example of Target Task Query>
</Target Task>

Please rewrite the Source Task Query to synthesize a new Target Task Query.

Output as follows:
<Rewrote>
<Target Task Query>
...
</Target Task Query>
</Rewrote>

(*@\textbf{ASSISTANT: }@*)
<Rewrote>
<Target Task Query>
Input:
Question: I struggled to breathe on the mountaintop.
 Option1: I got to an elevation that was higher.
 Option2: I got to an elevation that was lower.
Explanation:
</Target Task Query>
</Rewrote>
\end{lstlisting}\\\bottomrule
\end{tabular}
\caption{}
\end{table*}

\begin{table*}[hb] \small 
\centering
\begin{tabular}{p{\linewidth-1cm}}
\toprule
\textbf{Example of Prompt 2: Refine the synthesized query based on the user target task query.}\\
\midrule\begin{lstlisting}[escapeinside={(*@}{@*)}]
(*@\textbf{USER: }@*)
<Synthesized Query>
Input:
Question: I struggled to breathe on the mountaintop.
 Option1: I got to an elevation that was higher.
 Option2: I got to an elevation that was lower.
Explanation:
</Synthesized Query>

<Target Task>
<Target Task Instruction>
Given a premise and two alternatives, choose the alternative that is a more plausible cause or effect of the situation described by the premise. The input format is "premise
(1)alternative_1(2)alternative_2", the output should either be "1" or "2" based on your judgment.
</Target Task Instruction>
<Example of Target Task Query>
Input:
I ran out of breath.
(1)I climbed several flights of stairs.(2)I read several chapters of the book.

Give your final answer in the following format: "The final answer is: [your answer]"
</Example of Target Task Query>
</Target Task>

For the target task, I have synthesized a new query following the example of target task query. Could you help me refine the synthesized query?
Return your refined query in the following format:
<Refined Query>
...
</Refined Query>

(*@\textbf{ASSISTANT: }@*)
<Refined Query>
Input:
I struggled to breathe on the mountaintop.
(1)I got to an elevation that was higher.(2)I got to an elevation that was lower.

Give your final answer in the following format: "The final answer is: [your answer]"
</Refined Query>
\end{lstlisting}\\\bottomrule
\end{tabular}
\caption{}
\end{table*}

\begin{table*}[hb] \small 
\centering
\begin{tabular}{p{\linewidth-1cm}}
\toprule
\textbf{Example of Prompt 3: Generate the labels of the synthesized queries by referring to the original source tasks demonstration.}\\
\midrule\begin{lstlisting}[escapeinside={(*@}{@*)}]
(*@\textbf{USER: }@*)
Note: You may refer to the following source task demonstrations, which might help you solve the task.

Source Task Instruction:
You are given a question or fill-in-the-blank question, two answer options (Option1 and Option2) and an Explanation. Your task is to find the correct answer (return the string of the correct option, not option1/2) for the given question from the given options and using explanation.

Input:
Question: Dane was on a mountaintop and struggled to breathe, so he rectified it by getting to an elevation that was
 Option1: higher
 Option2: lower
 Explanation: At higher elevations, there is less air to press on a given area.

Give your final answer in the following format: "The final answer is: [your answer]"

The final answer is: lower



Please generate a response to the following target task question:

Target Task Instruction:
Given a premise and two alternatives, choose the alternative that is a more plausible cause or effect of the situation described by the premise. The input format is "premise
(1)alternative_1(2)alternative_2", the output should either be "1" or "2" based on your judgment.

Input:
I struggled to breathe on the mountaintop.
(1)I got to an elevation that was higher.(2)I got to an elevation that was lower.

Give your final answer in the following format: "The final answer is: [your answer]"

(*@\textbf{ASSISTANT: }@*)
The final answer is: 2
\end{lstlisting}\\\bottomrule
\end{tabular}
\caption{}
\end{table*}

\begin{table*}[hb] \small 
\centering
\begin{tabular}{p{\linewidth-1cm}}
\toprule
\textbf{Example of Prompt 4: Perform standard ICL using five source task demonstrations processed through our two-stage alignment strategy.}\\
\midrule\begin{lstlisting}[escapeinside={(*@}{@*)}]
(*@\textbf{USER: }@*)
Task Instruction:
Given a premise and two alternatives, choose the alternative that is a more plausible cause or effect of the situation described by the premise. The input format is "premise
(1)alternative_1(2)alternative_2", the output should either be "1" or "2" based on your judgment.

Input:
I struggled to breathe on the mountaintop.
(1)I got to an elevation that was higher.(2)I got to an elevation that was lower.

Give your final answer in the following format: "The final answer is: [your answer]"

The final answer is: 2

Task Instruction:
Given a premise and two alternatives, choose the alternative that is a more plausible cause or effect of the situation described by the premise. The input format is "premise
(1)alternative_1(2)alternative_2", the output should either be "1" or "2" based on your judgment.

Input:
The air pressure inside a patient's lungs decreases when the patient's chest gets larger.
(1)The patient is exhaling.(2)The patient is inhaling.

Give your final answer in the following format: "The final answer is: [your answer]"

The final answer is: 2

Task Instruction:
Given a premise and two alternatives, choose the alternative that is a more plausible cause or effect of the situation described by the premise. The input format is "premise
(1)alternative_1(2)alternative_2", the output should either be "1" or "2" based on your judgment.

Input:
The air pressure inside the lungs is lower than the air pressure outside.
(1)exhaling (2)inhaling

Give your final answer in the following format: "The final answer is: [your answer]"

The final answer is: 2

Task Instruction:
Given a premise and two alternatives, choose the alternative that is a more plausible cause or effect of the situation described by the premise. The input format is "premise
(1)alternative_1(2)alternative_2", the output should either be "1" or "2" based on your judgment.

Input:
Jeff is exercising his muscles at the gym. Over time his muscles will grow.
(1)bigger (2)smaller

Give your final answer in the following format: "The final answer is: [your answer]"

The final answer is: 1
\end{lstlisting}\\\midrule
\textbf{... Continue on next page}\\\bottomrule
\end{tabular}
\caption{}
\end{table*}

\begin{table*}[th] \small 
\centering
\begin{tabular}{p{\linewidth-1cm}}
\toprule
\textbf{... Continued from previous page}\\
\midrule\begin{lstlisting}[escapeinside={(*@}{@*)}]

Task Instruction:
Given a premise and two alternatives, choose the alternative that is a more plausible cause or effect of the situation described by the premise. The input format is "premise
(1)alternative_1(2)alternative_2", the output should either be "1" or "2" based on your judgment.

Input:
I inhale deeply.
(1)My air pressure increases.(2)My air pressure decreases.

Give your final answer in the following format: "The final answer is: [your answer]"

The final answer is: 1

Task Instruction:
Given a premise and two alternatives, choose the alternative that is a more plausible cause or effect of the situation described by the premise. The input format is "premise
(1)alternative_1(2)alternative_2", the output should either be "1" or "2" based on your judgment.

Input:
I ran out of breath.
(1)I climbed several flights of stairs.(2)I read several chapters of the book.

Give your final answer in the following format: "The final answer is: [your answer]"

(*@\textbf{ASSISTANT: }@*)
The final answer is: 1

\end{lstlisting}\\\bottomrule
\end{tabular}
\caption{}
\end{table*}

\clearpage
\section{Cross-Task Error Examples}
\label{appendix:errorexample}
This section presents examples of the various types of interference summarized in \S\ref{sec:crosstaskerroranalyze}. Please note that we omit some blank lines inside each question-answer pair to save space.

\captionsetup[table]{name=Error Example} 
\captionsetup{labelformat=empty} 
\setcounter{table}{0} 

\begin{table*}[h] \small 
\centering
\begin{tabular}{p{\linewidth-1cm}}
\toprule
\textbf{\textbf{Misleading Task Type}: The model responds to the target task input directly based on the source task description.}\\
\midrule\begin{lstlisting}[escapeinside={(*@}{@*)}]
(*@\textbf{USER: }@*)
Task Instruction:
In this task, you are given two sentences. Your task is to classify the given sentences as "Yes" if they have same meaning; otherwise, classify them as "No".
Input:
Sentence-1: I have a cat.<sep>Sentence-2: I have pets .
Give your final answer in the following format: "The final answer is: [your answer]"
The final answer is: Yes

Task Instruction:
In this task, you are given two sentences. Your task is to classify the given sentences as "Yes" if they have same meaning; otherwise, classify them as "No".
Input:
Sentence-1: I always eat meat<sep>Sentence-2: I can eat meat .
Give your final answer in the following format: "The final answer is: [your answer]"
The final answer is: Yes

Task Instruction:
In this task, you are given two sentences. Your task is to classify the given sentences as "Yes" if they have same meaning; otherwise, classify them as "No".
Input:
Sentence-1: There's a living room.<sep>Sentence-2: The flat does have an entertaiment space .
Give your final answer in the following format: "The final answer is: [your answer]"
The final answer is: Yes

Task Instruction:
In this task, you are given two sentences. Your task is to classify the given sentences as "Yes" if they have same meaning; otherwise, classify them as "No".
Input:
Sentence-1: I play piano<sep>Sentence-2: I play an instrument .
Give your final answer in the following format: "The final answer is: [your answer]"
The final answer is: Yes

Task Instruction:
In this task, you are given two sentences. Your task is to classify the given sentences as "Yes" if they have same meaning; otherwise, classify them as "No".
Input:
Sentence-1: if there's a new club in town.<sep>Sentence-2: Night night .
Give your final answer in the following format: "The final answer is: [your answer]"
The final answer is: Yes

Task Instruction:
In this task, you're given a pair of sentences, sentence 1 and sentence 2. Your job is to choose whether the two sentences clearly agree (entailment)/disagree (contradiction) with each other, or if this cannot be determined (neutral). Your answer must be in the form of the letters E, C, and N respectively.
Input:
Sentence 1: A woman wearing a fur coat enters through a doorway Sentence 2: The woman enters a club.
Give your final answer in the following format: "The final answer is: [your answer]"

(*@\textbf{ASSISTANT: }@*)
The final answer is: No
\end{lstlisting}\\\bottomrule
\end{tabular}
\caption{}
\end{table*}

\begin{table*}[h] \small 
\centering
\begin{tabular}{p{\linewidth-1cm}}
\toprule
\textbf{\textbf{Confusing Output Format}: When the output formats of the target and source tasks are similar, the model may replicate the output format of source tasks, including label space, output constraints (e.g., capitalization), imitation of text style and length, etc.}\\
\midrule\begin{lstlisting}[escapeinside={(*@}{@*)}]
(*@\textbf{USER: }@*)
Task Instruction:
In this task, you are given a context and four options. Each option is a suggested ending for the context. You should read the context and pick the best ending for the context. Please answer with "A", "B", "C", and "D".
Input:
A woman is standing outside in the snow talking to a man. he <sep> (A) puts a fire in a pit of wood and lights a match. (B) helps her snowboard back into the car and they drive away. (C) is holding a black snow shovel. (D) is lighting a fire while she watches.
Give your final answer in the following format: "The final answer is: [your answer]"
The final answer is: C

Task Instruction:
In this task, you are given a context and four options. Each option is a suggested ending for the context. You should read the context and pick the best ending for the context. Please answer with "A", "B", "C", and "D".
Input:
Two bodybuilder women are seated at a table. they <sep> (A) are talking about diving techniques, bribing each other with muscle' n strength. (B) are working out on exercise bikes. (C) are arm wrestling, vieing to win. (D) are shown on parallel bars.
Give your final answer in the following format: "The final answer is: [your answer]"
The final answer is: C

Task Instruction:
In this task, you are given a context and four options. Each option is a suggested ending for the context. You should read the context and pick the best ending for the context. Please answer with "A", "B", "C", and "D".
Input:
A man is talking outside a house. he <sep> (A) is using a scraper to remove rocks. (B) opens a large box and tools are inside. (C) is discussing the tree in the yard. (D) carries a tire on his shoulders.
Give your final answer in the following format: "The final answer is: [your answer]"
The final answer is: C

Task Instruction:
In this task, you are given a context and four options. Each option is a suggested ending for the context. You should read the context and pick the best ending for the context. Please answer with "A", "B", "C", and "D".
Input:
A young girl stands in the drive way. She starts backing up toward the street. she <sep> (A) then reaches an intersection and turns right. (B) then runs forward toward the camera. (C) looked at her phone. (D) then gracefully bends and slides down the sidewalk.
Give your final answer in the following format: "The final answer is: [your answer]"
The final answer is: B

Task Instruction:
In this task, you are given a context and four options. Each option is a suggested ending for the context. You should read the context and pick the best ending for the context. Please answer with "A", "B", "C", and "D".
Input:
A pitcher pitches a ball to the batter. the batter <sep> (A) bats the ball really far. (B) is deflected by the bat. (C) misses, throwing the ball into the field. (D) takes the envelope down and lays it down.
Give your final answer in the following format: "The final answer is: [your answer]"
The final answer is: A

Task Instruction:
In this task, you are given a question containing a blank (_) and two options. You should pick the best option to answer the question. Please answer with "A" or "B".
Input:
Mary wanted to buy a house from Natalie, but _ could not pay market value. (A) Mary (B) Natalie
Give your final answer in the following format: "The final answer is: [your answer]"

(*@\textbf{ASSISTANT: }@*)
The final answer is: C

\end{lstlisting}\\\bottomrule
\end{tabular}
\caption{}
\end{table*}

\begin{table*}[h] \small 
\centering
\begin{tabular}{p{\linewidth-1cm}}
\toprule
\textbf{\textbf{Keyword Shift}: Same keywords in both tasks may have different meanings. The model may mistakenly interpret the keyword based on the source task.}\\
\midrule\begin{lstlisting}[escapeinside={(*@}{@*)}]
(*@\textbf{USER: }@*)
Task Instruction:
You're given a sentence and your task is to classify whether the sentence is acceptable or not. Any sentence which is grammatically correct, has a naturalistic text is acceptable, otherwise unacceptable.
Input:
I doubt if you can help me in understanding this.
Give your final answer in the following format: "The final answer is: [your answer]"
The final answer is: acceptable

Task Instruction:
You're given a sentence and your task is to classify whether the sentence is acceptable or not. Any sentence which is grammatically correct, has a naturalistic text is acceptable, otherwise unacceptable.
Input:
I doubt if you to help me in understanding this.
Give your final answer in the following format: "The final answer is: [your answer]"
The final answer is: unacceptable

Task Instruction:
You're given a sentence and your task is to classify whether the sentence is acceptable or not. Any sentence which is grammatically correct, has a naturalistic text is acceptable, otherwise unacceptable.
Input:
This is the problem that you'll beat the competition more easily, the sooner you solve.
Give your final answer in the following format: "The final answer is: [your answer]"
The final answer is: unacceptable

Task Instruction:
You're given a sentence and your task is to classify whether the sentence is acceptable or not. Any sentence which is grammatically correct, has a naturalistic text is acceptable, otherwise unacceptable.
Input:
That's the kindest answer that I ever heard.
Give your final answer in the following format: "The final answer is: [your answer]"
The final answer is: acceptable

Task Instruction:
You're given a sentence and your task is to classify whether the sentence is acceptable or not. Any sentence which is grammatically correct, has a naturalistic text is acceptable, otherwise unacceptable.
Input:
All native speakers have a grammatical competence which can generate an infinite set of grammatical sentences from a finite set of resources.
Give your final answer in the following format: "The final answer is: [your answer]"
The final answer is: acceptable

Task Instruction:
Given the Question and Answer pair, detect whether the answer is acceptable or not. Denote acceptable answer as "yes", otherwise denote as "no". An acceptable answer is consistent with the context in question, detailed, long, and uses general knowledge of English. Answers that are not definite and suggest a variety of options are not acceptable answers.
Input:
Question: I must admit that I'm unsure whether or not this is the right forum for this question. It may belong to statistics and AI also. If there exists a more suitable forum, then tell me. The thing is, I want to analyse a lot of English text for an AI project (Confabulation theory). Is there an online collection of freely available English texts? Books, news would be preferred scientific texts will properly not do, due to large amounts of math etc.
Answer: And for modern texts, there are a lot of blogs which use various Creative Commons licenses, most commonly CC BY-SA.
Give your final answer in the following format: "The final answer is: [your answer]"

(*@\textbf{ASSISTANT: }@*)
The final answer is: yes
\end{lstlisting}\\\bottomrule
\end{tabular}
\caption{}
\end{table*}

\begin{table*}[h] \small 
\centering
\begin{tabular}{p{\linewidth-1cm}}
\toprule
\textbf{\textbf{Label Distribution Shift}: When source and target task output labels share similar semantics, the label distribution of the former may bias the latter.}\\
\midrule\begin{lstlisting}[escapeinside={(*@}{@*)}]
(*@\textbf{USER: }@*)
Task Instruction:
In this task, you are given a question and an answer. Answer "Yes" if the given answer correctly answers the question, otherwise answer "No".
Input:
how much is 1 tablespoon of water, Answer: The capacity of ordinary tablespoons is not regulated by law and is subject to considerable variation.
Give your final answer in the following format: "The final answer is: [your answer]"
The final answer is: No

Task Instruction:
In this task, you are given a question and an answer. Answer "Yes" if the given answer correctly answers the question, otherwise answer "No".
Input:
how are the # of electrons in each shell determined, Answer: (See Madelung rule for more details.)
Give your final answer in the following format: "The final answer is: [your answer]"
The final answer is: No

Task Instruction:
In this task, you are given a question and an answer. Answer "Yes" if the given answer correctly answers the question, otherwise answer "No".
Input:
how long is the term for federal judges, Answer: See Article I and Article III tribunals .
Give your final answer in the following format: "The final answer is: [your answer]"
The final answer is: No

Task Instruction:
In this task, you are given a question and an answer. Answer "Yes" if the given answer correctly answers the question, otherwise answer "No".
Input:
how much is 1 tablespoon of water, Answer: It is abbreviated as T, tb, tbs, tbsp, tblsp, or tblspn.
Give your final answer in the following format: "The final answer is: [your answer]"
The final answer is: No

Task Instruction:
In this task, you are given a question and an answer. Answer "Yes" if the given answer correctly answers the question, otherwise answer "No".
Input:
How Do You Find the mean of the squares of the first 10 counting numbers, Answer: The concept of square can be extended to some other number systems.
Give your final answer in the following format: "The final answer is: [your answer]"
The final answer is: No

Task Instruction:
The answer will be 'yes' if the provided sentence contains an explicit mention that answers the given question. Otherwise, the answer should be 'no'. Instances where the answer is implied from the sentence using "instinct" or "common sense" (as opposed to being written explicitly in the sentence) should be labeled as 'no'.
Input:
Sentence: So from now on, Marsha makes sure Joey is safe and sound at all times.
Question: How often was Joey safe and sound?
Give your final answer in the following format: "The final answer is: [your answer]"

(*@\textbf{ASSISTANT: }@*)
The final answer is: no
\end{lstlisting}\\\bottomrule
\end{tabular}
\caption{}
\end{table*}

\begin{table*}[h] \small 
\centering
\begin{tabular}{p{\linewidth-1cm}}
\toprule
\textbf{\textbf{Simplified Thinking}: If the source task is simple, LLMs may approach the complex target task in an overly direct manner.}\\
\midrule\begin{lstlisting}[escapeinside={(*@}{@*)}]
(*@\textbf{USER: }@*)

Task Instruction:
You are given a sentence and a question, construct 2 answer options in a specific format i.e. ['option1','option2'].
Input:
Sentence: The man lifted the boy onto his bunk bed. Question: Whose bunk bed?
Give your final answer in the following format: "The final answer is: [your answer]"
The final answer is: ["boy's", "man's"]

Task Instruction:
You are given a sentence and a question, construct 2 answer options in a specific format i.e. ['option1','option2'].
Input:
Sentence: Stretching her back, the woman smiled at the girl. Question: Whose back was the woman stretching?
Give your final answer in the following format: "The final answer is: [your answer]"
The final answer is: ["girl's", "woman's"]

Task Instruction:
You are given a sentence and a question, construct 2 answer options in a specific format i.e. ['option1','option2'].
Input:
Sentence: Bill passed the gameboy to John because his turn was over. Question: Whose turn was over?
Give your final answer in the following format: "The final answer is: [your answer]"
The final answer is: ["Bill's", "John's"]

Task Instruction:
You are given a sentence and a question, construct 2 answer options in a specific format i.e. ['option1','option2'].
Input:
Sentence: Bill passed the gameboy to John because his turn was next. Question: Whose turn was next?
Give your final answer in the following format: "The final answer is: [your answer]"
The final answer is: ["Bill's", "John's"]

Task Instruction:
You are given a sentence and a question, construct 2 answer options in a specific format i.e. ['option1','option2'].
Input:
Sentence: Patting her back, the woman smiled at the girl. Question: Whose back was the woman patting?
Give your final answer in the following format: "The final answer is: [your answer]"
The final answer is: ["girl's", "woman's"]

Task Instruction:
In this task you need to indicate the plausibility of reasoning for the pronoun coreference relations. Each of the provided inputs contains a sentence with a target pronoun and a sentence that justifies which noun phrase the pronoun refers to. Correct reasons do not need to use all the knowledge from the sentence. The resolution of the pronoun coreference relations typically involve one or multiple following knowledge types about commonsense: First: 'Property', the knowledge about property of objects (e.g., ice is cold). Second: 'Object', the knowledge about objects (e.g., cats have ears). Third: 'Eventuality', the knowledge about eventuality (e.g., 'wake up' happens before 'open eyes'). Forth: 'Spatial', the knowledge about spatial position (e.g., object at the back can be blocked). Fifth: 'Quantity', the knowledge about numbers (e.g., 2 is smaller than 10). Sixth: all other knowledge if above ones are not suitable. You should answer 'Correct' if the reasoning made sense, otherwise, you should answer 'Wrong'.
Input:
Sentence: Many people start to read Paul's books and can't put them down. They are gripped because Paul writes so well.
Reason: The 'They' refers to people because People love Paul's books. They are popular. So the books are popular, not the people.
Question: Is the above reasoning correct or wrong?
Give your final answer in the following format: "The final answer is: [your answer]"

(*@\textbf{ASSISTANT: }@*)
The final answer is: Correct
\end{lstlisting}\\\bottomrule
\end{tabular}
\caption{}
\end{table*}

\end{document}